%% file: paper.tex
\documentclass[10pt,twocolumn,letterpaper]{article}
\newif\ifforceplain\forceplainfalse %

\usepackage{cvpr}					%

\newif\ifarxiv
\arxivtrue
\ifarxiv\cvprfinalcopy\fi

\makeatletter						%
\@namedef{ver@everyshi.sty}{}
\makeatother
\usepackage{tikz}

\usepackage{times}					%

\usepackage{epsfig}
\usepackage{graphicx}
\usepackage{amsmath}
\usepackage{amssymb}
\usepackage{array}
\usepackage{multirow}
\usepackage{fancyhdr}

\usepackage[mathscr]{eucal}
\usepackage{amsfonts}
\usepackage{color}
\usepackage[normalem]{ulem}  %

\newcommand{\ext}{jpg}  %
\newcommand{\extt}{jpg} %

\newcommand{\extraspaceafter}{}
\newif\iflualatex
\lualatexfalse
\iflualatex
	\renewcommand{\extraspaceafter}{\vspace*{1.5mm}}
	\renewcommand{\extt}{jp2}
	\usepackage[no-math]{fontspec}		%
\fi

\usepackage[pagebackref=true,breaklinks=true,letterpaper=true,colorlinks,bookmarks=false]{hyperref}

\def\cvprPaperID{2860} %

\newcommand{\arch}[1]{\textsc{#1}}

\DeclareMathOperator{\EX}{\mathbb{E}}%

\def\clap#1{\hbox to 0pt{\hss #1\hss}}%
\definecolor{internationalkleinblue}{rgb}{0.0, 0.18, 0.65}	%
\definecolor{olive}{rgb}{0.5, 0.5, 0.0}
\definecolor{gray}{rgb}{0.7, 0.7, 0.7}
\definecolor{maroon}{rgb}{0.69, 0.19, 0.38}
\definecolor{celestialblue}{rgb}{0.29, 0.59, 0.82}
\definecolor{red}{rgb}{1.00, 0.00, 0.00}
\newcommand{\slerp}{\mathrm{slerp}}
\newcommand{\lerp}{\mathrm{lerp}}

\newcommand{\FINAL}[2][]{#2} %

\newcommand{\xx}{{\bf x}}
\newcommand{\yy}{{\bf y}}
\newcommand{\zz}{{\bf z}}
\newcommand{\ww}{{\bf w}}
\newcommand{\barww}{{\bf\bar w}}

\newcommand{\ZZ}{\mathcal{Z}}
\newcommand{\WW}{\mathcal{W}}

\newcommand{\codelink}{\FINAL{\texttt{https://github.com/NVlabs/stylegan}}}
\newcommand{\ffhqlink}{\FINAL{\texttt{\small{https://github.com/NVlabs/ffhq-dataset}}}}

\input{figures}

\ifcvprfinal\fi
\begin{document}

\title{A Style-Based Generator Architecture for Generative Adversarial Networks}

\author{Tero Karras\\
NVIDIA\\
{\tt\small tkarras@nvidia.com}
\and
Samuli Laine\\
NVIDIA\\
{\tt\small slaine@nvidia.com}
\and
Timo Aila\\
NVIDIA\\
{\tt\small taila@nvidia.com}
}

\maketitle

\ifcvprfinal\thispagestyle{empty}\fi
\ifarxiv
\pagestyle{plain}\thispagestyle{plain}
\fi

\ifforceplain\thispagestyle{plain}\pagestyle{plain}\fi

\begin{abstract}

We propose an alternative generator architecture for generative adversarial networks, borrowing from style transfer literature. The new architecture leads to an automatically learned, unsupervised separation of high-level attributes (e.g.,~pose and identity when trained on human faces) and stochastic variation in the generated images (e.g.,~freckles, hair), and it enables intuitive, scale-specific control of the synthesis. The new generator improves the state-of-the-art in terms of traditional distribution quality metrics, leads to demonstrably better interpolation properties, and also better disentangles the latent factors of variation. To quantify interpolation quality and disentanglement, we propose two new, automated methods that are applicable to any generator architecture. Finally, we introduce a new, highly varied and high-quality dataset of human faces. %
\extraspaceafter\extraspaceafter
\end{abstract}

\section{Introduction}

The resolution and quality of images produced by generative methods\,---\,especially generative adversarial networks (GAN) \cite{Goodfellow2014}\,---\,have seen rapid improvement recently \cite{Karras2017,Miyato2018B,Brock2018}. Yet the generators continue to operate as black boxes, and despite recent efforts \cite{JunYan2018}, the understanding of various aspects of the image synthesis process, e.g., the origin of stochastic features, is still lacking. The properties of the latent space are also poorly understood, and the commonly demonstrated latent space interpolations \cite{Dosovitskiy2015,Sainburg2018,Laine2018iclr} provide no quantitative way to compare different generators against each other.

Motivated by style transfer literature \cite{Huang2017}, we re-design the generator architecture in a way that exposes novel ways to control the image synthesis process. Our generator starts from a learned constant input and adjusts the ``style'' of the image at each convolution layer based on the latent code, therefore directly controlling the strength of image features at different scales. Combined with noise injected directly into the network, this architectural change leads to automatic, unsupervised separation of high-level attributes (e.g.,~pose, identity) from stochastic variation (e.g.,~freckles, hair) in the generated images, and enables intuitive scale-specific mixing and interpolation operations. We do not modify the discriminator or the loss function in any way, and our work is thus orthogonal to the ongoing discussion about GAN loss functions, regularization, and hyper-parameters \cite{Gulrajani2017,Miyato2018B,Brock2018,Lucic2017,Mescheder2018,Kurach2018}.

Our generator embeds the input latent code into an intermediate latent space, which has a profound effect on how the factors of variation are represented in the network. The input latent space must follow the probability density of the training data, and we argue that this leads to some degree of unavoidable entanglement. %
Our intermediate latent space is free from that restriction and is therefore allowed to be disentangled. As previous methods for estimating the degree of latent space disentanglement are not directly applicable in our case, we propose two new automated metrics\,---\,perceptual path length and linear separability\,---\,for quantifying these aspects of the generator. Using these metrics, we show that compared to a traditional generator architecture, our generator admits a more linear, less entangled representation of different factors of variation.

Finally, we present a new dataset of human faces (Flickr-Faces-HQ, FFHQ) that offers much higher quality and covers considerably wider variation than existing high-resolution datasets (Appendix~\ref{sec:FFHQ}). 
We \FINAL{have made} this dataset publicly available, along with our source code and pre-trained networks.\footnote{\codelink}
The accompanying video can be found \FINAL{under the same link}.

\section{Style-based generator}
\label{sec:arch}

\figarch

Traditionally the latent code is provided to the generator through an input layer, i.e.,~the first layer of a feedforward network (Figure~\ref{fig:arch}a). We depart from this design by omitting the input layer altogether and starting from a learned constant instead (Figure~\ref{fig:arch}b, right). Given a latent code $\zz$ in the input latent space $\ZZ$, a non-linear mapping network \mbox{$f:\ZZ\to\WW$} first produces $\ww \in \WW$ (Figure~\ref{fig:arch}b, left). For simplicity, we set the dimensionality of both spaces to 512, and the mapping $f$ is implemented using an 8-layer MLP, a decision we will analyze in Section~\ref{sec:pathlen}.
Learned affine transformations then specialize $\ww$ to \emph{styles} $\yy = (\yy_s, \yy_b)$ that control adaptive instance normalization \FINAL{(AdaIN)~\cite{Huang2017,Dumoulin2016,Ghiasi2017,Dumoulin2018}} operations after each convolution layer of the synthesis network $g$. The AdaIN operation is defined as
\begin{equation}
\label{eqn:adain}
	\textrm{AdaIN}(\xx_i,\yy) = \yy_{s,i}\frac{\xx_i-\mu(\xx_i)}{\sigma(\xx_i)} + \yy_{b,i}	\textrm{,}	%
\end{equation}
where each feature map $\xx_i$ is normalized separately, and then scaled and biased using the corresponding scalar components from style $\yy$. Thus the dimensionality of $\yy$ is twice the number of feature maps on that layer.

Comparing our approach to style transfer, we compute the spatially invariant style $\yy$ from vector $\ww$ instead of an example image. We choose to reuse the word ``style'' for $\yy$ because similar network architectures are already used for feedforward style transfer \cite{Huang2017}, unsupervised image-to-image translation \cite{Huang2018}, and domain mixtures \cite{Hao2018}.\FINAL{ Compared to more general feature transforms~\cite{Li2017C,Siarohin2018}, AdaIN is particularly well suited for our purposes due to its efficiency and compact representation.}

Finally, we provide our generator with a direct means to generate stochastic detail by introducing explicit \emph{noise inputs}. These are single-channel images consisting of uncorrelated Gaussian noise, and we feed a dedicated noise image to each layer of the synthesis network. The noise image is broadcasted to all feature maps using learned per-feature scaling factors and then added to the output of the corresponding convolution, as illustrated in Figure~\ref{fig:arch}b. The implications of adding the noise inputs are discussed in Sections~\ref{sec:stoch_variation} and~\ref{sec:stoch_separation}.

\tabFID

\subsection{Quality of generated images}
Before studying the properties of our generator, we demonstrate experimentally that the redesign does not compromise image quality but, in fact, improves it considerably. Table~\ref{tab:FID} gives Fr\'echet inception distances (FID) \cite{Heusel2017} for various generator architectures in \textsc{CelebA-HQ} \cite{Karras2017} and our new \textsc{FFHQ} dataset (Appendix~\ref{sec:FFHQ}). 
\ifarxiv
Results for other datasets are given in Appendix~\ref{sec:otherdatasets}.
\else
Results for other datasets are given in the supplement.
\fi
Our baseline configuration (\arch{a}) is the Progressive GAN setup of Karras~et~al.~\cite{Karras2017}, from which we inherit the networks and all hyperparameters except where stated otherwise. We first switch to an improved baseline (\arch{b}) by using bilinear up/downsampling \FINAL{operations \cite{zhang2019}}, longer training, and tuned hyperparameters. 
\ifarxiv
A detailed description of training setups and hyperparameters is included in Appendix~\ref{sec:hyperparams}.
\else
A detailed description of training setups and hyperparameters is included in the supplement.
\fi
We then improve this new baseline further by adding the mapping network and AdaIN operations (\arch{c}), and make a surprising observation that the network no longer benefits from feeding the latent code into the first convolution layer. We therefore simplify the architecture by removing the traditional input layer and starting the image synthesis from a learned \mbox{$4\times4\times512$} constant tensor (\arch{d}). We find it quite remarkable that the synthesis network is able to produce meaningful results even though it receives input only through the styles that control the AdaIN operations.

Finally, we introduce the noise inputs (\arch{e}) that improve the results further, as well as novel \emph{mixing regularization} (\arch{f}) that decorrelates neighboring styles and enables more fine-grained control over the generated imagery (Section~\ref{sec:style_mixing}).

\figquality

We evaluate our methods using two different loss functions: for \textsc{CelebA-HQ} we rely on WGAN-GP \cite{Gulrajani2017}, while \textsc{FFHQ} uses WGAN-GP for configuration \arch{a} and non-saturating loss \cite{Goodfellow2014} with $R_1$ regularization \cite{Mescheder2018,Ross2017,Drucker1992} for configurations \arch{b}--\arch{f}. We found these choices to give the best results. Our contributions do not modify the loss function.%

We observe that the style-based generator (\arch{e}) improves FIDs quite significantly over the traditional generator (\arch{b}), almost 20\%, corroborating the large-scale ImageNet measurements made in parallel work~\cite{Chen2018self,Brock2018}. %
Figure~\ref{fig:quality} shows an uncurated set of novel images generated from the FFHQ dataset using our generator. As confirmed by the FIDs, the average quality is high, and even accessories such as eyeglasses and hats get successfully synthesized. For this figure, we avoided sampling from the extreme regions of $\WW$ using the so-called truncation \FINAL{trick~\cite{Marchesi2017, Brock2018, Kingma2018}}\,---\,Appendix~\ref{sec:truncation} details how the trick can be performed in $\WW$ instead of $\ZZ$. Note that our generator allows applying the truncation selectively to low resolutions only, so that high-resolution details are not affected.

All FIDs in this paper are computed without the truncation trick, and we only use it for illustrative purposes in Figure~\ref{fig:quality} and the video. All images are generated in $1024^2$ resolution.

\subsection{Prior art}
\label{sec:prior_art}
Much of the work on GAN architectures has focused on improving the discriminator by, e.g., using multiple discriminators \cite{Durugkar2016,Mordido2018,Doan2018}, multiresolution discrimination \cite{Wang2017,Sharma2018}, or self-attention \cite{Zhang2018sagan}. The work on generator side has mostly focused on the exact distribution in the input latent space \cite{Brock2018} or
shaping the input latent space via Gaussian mixture models \cite{BenYosef2018}, clustering \cite{Mukherjee2018}, or encouraging convexity \cite{Sainburg2018}.

Recent conditional generators feed the class identifier through a separate embedding network to a large number of layers in the generator \cite{Miyato2018}, while the latent is still provided though the input layer.
A few authors have considered feeding parts of the latent code to multiple generator layers \cite{Denton2015,Brock2018}. In parallel work, Chen et al.~\cite{Chen2018self} ``self modulate'' the generator using AdaINs, similarly to our work, but do not consider an intermediate latent space or noise inputs.

\figstylemix

\section{Properties of the style-based generator}
\label{sec:properties}
\label{sec:localized_styles}

Our generator architecture makes it possible to control the image synthesis via scale-specific modifications to the styles. We can view the mapping network and affine transformations as a way to draw samples for each style from a learned distribution, and the synthesis network as a way to generate a novel image based on a collection of styles. The effects of each style are localized in the network, i.e., modifying a specific subset of the styles can be expected to affect only certain aspects of the image.

To see the reason for this localization, let us consider how the AdaIN operation (Eq.~\ref{eqn:adain}) first normalizes each channel to zero mean and unit variance, and only then applies scales and biases based on the style. The new per-channel statistics, as dictated by the style, modify the relative importance of features for the subsequent convolution operation, but they do not depend on the original statistics because of the normalization. Thus each style controls only one convolution before being overridden by the next AdaIN operation.

\subsection{Style mixing}
\label{sec:style_mixing}

To further encourage the styles to localize, we employ \emph{mixing regularization}, where a given percentage of images are generated using two random latent codes instead of one during training. When generating such an image, we simply switch from one latent code to another\,---\,an operation we refer to as {\em style mixing}\,---\,at a randomly selected point in the synthesis network. To be specific, we run two latent codes $\zz_1, \zz_2$ through the mapping network, and have the corresponding $\ww_1, \ww_2$ control the styles so that $\ww_1$ applies before the crossover point and $\ww_2$ after it. This regularization technique prevents the network from assuming that adjacent styles are correlated.

Table~\ref{tab:mixing} shows how enabling mixing regularization during training improves the localization considerably, indicated by improved FIDs in scenarios where multiple latents are mixed at test time. Figure~\ref{fig:stylemix} presents examples of images synthesized by mixing two latent codes at various scales. We can see that each subset of styles controls meaningful high-level attributes of the image.

\tabmixing

\subsection{Stochastic variation}
\label{sec:stoch_variation}

\fignoisedetail

There are many aspects in human portraits that can be regarded as stochastic, such as the exact placement of hairs, stubble, freckles, or skin pores. Any of these can be randomized without affecting our perception of the image as long as they follow the correct distribution.

Let us consider how a traditional generator implements stochastic variation. Given that the only input to the network is through the input layer, the network needs to invent a way to generate spatially-varying pseudorandom numbers from earlier activations whenever they are needed. This consumes network capacity and hiding the periodicity of generated signal is difficult\,---\,and not always successful, as evidenced by commonly seen repetitive patterns in generated images. Our architecture sidesteps these issues altogether by adding per-pixel noise after each convolution.

Figure~\ref{fig:noisedetail} shows stochastic realizations of the same underlying image, produced using our generator with different noise realizations. We can see that the noise affects only the stochastic aspects, leaving the overall composition and high-level aspects such as identity intact. Figure~\ref{fig:noisemix} further illustrates the effect of applying stochastic variation to different subsets of layers. Since these effects are best seen in animation, please consult the accompanying video for a demonstration of how changing the noise input of one layer leads to stochastic variation at a matching scale. 

\fignoisemix

We find it interesting that the effect of noise appears tightly localized in the network. 
We hypothesize that at any point in the generator, there is pressure to introduce new content as soon as possible, and the easiest way for our network to create stochastic variation is to rely on the noise provided. A fresh set of noise is available for every layer, and thus there is no incentive to generate the stochastic effects from earlier activations, leading to a localized effect.

\subsection{Separation of global effects from stochasticity}
\label{sec:stoch_separation}

The previous sections as well as the accompanying video demonstrate that while changes to the style have global effects (changing pose, identity, etc.), the noise affects only inconsequential stochastic variation (differently combed hair, beard, etc.). This observation is in line with style transfer literature, where it has been established that spatially invariant statistics (Gram matrix, channel-wise mean, variance, etc.) reliably encode the style of an image \cite{Gatys2016,Li2017B} while spatially varying features encode a specific instance. 

In our style-based generator, the style affects the entire image because complete feature maps are scaled and biased with the same values. Therefore, global effects such as pose, lighting, or background style can be controlled coherently. Meanwhile, the noise is added independently to each pixel and is thus ideally suited for controlling stochastic variation. If the network tried to control, e.g., pose using the noise, that would lead to spatially inconsistent decisions that would then be penalized by the discriminator. Thus the network learns to use the global and local channels appropriately, without explicit guidance.

\section{Disentanglement studies}
\label{sec:disentanglement}

\figillustration

There are various definitions for disentanglement \cite{Schmidhuber92,Ridgeway2016,Achille2017,Chen2018,Eastwood2018}, but a common goal is a latent space that consists of linear subspaces, each of which controls one factor of variation.
However, the sampling probability of each combination of factors in $\ZZ$ needs to match the corresponding density in the training data. 
As illustrated in Figure~\ref{fig:illustration}, 
this precludes the factors from being fully disentangled with typical datasets and input latent distributions.%
\footnote{The few artificial datasets designed for disentanglement studies (e.g., \cite{dsprites,Eastwood2018}) tabulate all combinations of predetermined factors of variation with uniform frequency, thus hiding the problem.}

A major benefit of our generator architecture is that the intermediate latent space $\WW$ does not have to support sampling according to any \emph{fixed} distribution; its sampling density is induced by the \emph{learned} piecewise continuous mapping $f(\zz)$. 
This mapping can be adapted to ``unwarp'' $\WW$ so that the factors of variation become more linear. 
We posit that there is pressure for the generator to do so, as it should be easier to generate realistic images based on a disentangled representation than based on an entangled representation.
As such, we expect the training to yield a less entangled $\WW$ in an unsupervised setting, i.e., when the factors of variation are not known in advance \cite{Desjardins2012,Kingma2014VAE,Rezende2014,Chen2016,Higgins2017,Kim2018,Chen2018}.

Unfortunately the metrics recently proposed for quantifying disentanglement \cite{Higgins2017,Kim2018,Chen2018,Eastwood2018} require an encoder network that maps input images to latent codes. These metrics are ill-suited for our purposes since our baseline GAN lacks such an encoder. While it is possible to add an extra network for this purpose \cite{Chen2016,Donahue2016,Dumoulin2017}, we want to avoid investing effort into a component that is not a part of the actual  solution. To this end, we describe two new ways of quantifying disentanglement, neither of which requires an encoder or known factors of variation, and are therefore computable for any image dataset and generator.

\subsection{Perceptual path length}
\label{sec:pathlen}

As noted by Laine~\cite{Laine2018iclr}, interpolation of latent-space vectors may yield surprisingly non-linear changes in the image. For example, features that are absent in either endpoint may appear in the middle of a linear interpolation path. This is a sign that the latent space is entangled and the factors of variation are not properly separated. To quantify this effect, we can measure how drastic changes the image undergoes as we perform interpolation in the latent space. Intuitively, a less curved latent space should result in perceptually smoother transition than a highly curved latent space.

As a basis for our metric, we use a perceptually-based pairwise image distance~\cite{Zhang2018metric} that is calculated as a weighted difference between two VGG16~\cite{simonyan2014} embeddings, where the weights are fit so that the metric agrees with human perceptual similarity judgments. If we subdivide a latent space interpolation path into linear segments, we can define the total perceptual length of this segmented path as the sum of perceptual differences over each segment, as reported by the image distance metric. A natural definition for the perceptual path length would be the limit of this sum under infinitely fine subdivision, but in practice we approximate it using a small subdivision epsilon $\epsilon=10^{-4}$. The average perceptual path length in latent space $\ZZ$, over all possible endpoints, is therefore
\begin{equation}
\begin{array}{r@{}l}
\hspace*{-3mm}l_{\ZZ} = \EX\Big[{\displaystyle\frac{1}{\epsilon^2}}d\big(&G(\slerp(\zz_1,\zz_2;\,t)), \\
&G(\slerp(\zz_1,\zz_2;\,t+\epsilon))\big)\Big]\textrm{,}
\end{array}
\end{equation}
where \mbox{$\zz_1,\zz_2\sim P(\zz),t\sim U(0,1)$}, $G$ is the generator (i.e., $g \circ f$ for style-based networks), and $d(\cdot,\cdot)$ evaluates the perceptual distance between the resulting images. Here $\slerp$ denotes spherical interpolation~\cite{shoemake85}, which is the most appropriate way of interpolating in our normalized input latent space~\cite{white16}. To concentrate on the facial features instead of background, we crop the generated images to contain only the face prior to evaluating the pairwise image metric. 
As the metric $d$ is quadratic~\cite{Zhang2018metric}, we divide by $\epsilon^2$.
We compute the expectation by taking 100,000 samples.

\figFFHQ

Computing the average perceptual path length in $\WW$ is carried out in a similar fashion:
\begin{equation}
\begin{array}{r@{}l}
\hspace*{-3mm}l_{\WW} = \EX\Big[{\displaystyle\frac{1}{\epsilon^2}}d\big(&g(\lerp(f(\zz_1),f(\zz_2);\,t)), \\
&g(\lerp(f(\zz_1),f(\zz_2);\,t+\epsilon))\big)\Big]\textrm{,}
\end{array}
\end{equation}
where the only difference is that interpolation happens in $\WW$ space. Because vectors in $\WW$ are not normalized in any fashion, we use linear interpolation ($\lerp$).

\tabdisentangle

Table~\ref{tab:disentangle} shows that this full-path length is substantially shorter for our style-based generator with noise inputs, indicating that $\WW$ is perceptually more linear than $\ZZ$.
Yet, this measurement is in fact slightly biased in favor of the input latent space $\ZZ$. If $\WW$ is indeed a disentangled and ``flattened'' mapping of $\ZZ$, it may contain regions that are not on the input manifold\,---\,and are thus badly reconstructed by the generator\,---\,even between points that are mapped from the input manifold, whereas the input latent space $\ZZ$ has no such regions by definition. It is therefore to be expected that if we restrict our measure to path endpoints, i.e., $t \in \{0,1\}$, we should obtain a smaller $l_{\WW}$ while $l_{\ZZ}$ is not affected. This is indeed what we observe in Table~\ref{tab:disentangle}.

Table~\ref{tab:mapping} shows how path lengths are affected by the mapping network. %
We see that both traditional and style-based generators benefit from having a mapping network, and additional depth generally improves the perceptual path length as well as FIDs.
It is interesting that while $l_{\WW}$ improves in the traditional generator, $l_{\ZZ}$ becomes considerably worse, illustrating our claim that the input latent space can indeed be arbitrarily entangled in GANs.

\tabmapping

\subsection{Linear separability}
\label{sec:separability}

If a latent space is sufficiently disentangled, it should be possible to find direction vectors that consistently correspond to individual factors of variation. We propose another metric that quantifies this effect by measuring how well the latent-space points can be separated into two distinct sets via a linear hyperplane, so that each set corresponds to a specific binary attribute of the image.

In order to label the generated images, we train auxiliary classification networks for a number of binary attributes, e.g., to distinguish male and female faces. 
In our tests, the classifiers had the same architecture as the discriminator we use (i.e., same as in \cite{Karras2017}), and were trained using the \textsc{CelebA-HQ} dataset that retains the~40 attributes available in the original CelebA dataset.
To measure the separability of one attribute, we generate 200,000 images with $\zz\sim P(\zz)$ and classify them using the auxiliary classification network. 
We then sort the samples according to classifier confidence and remove the least confident half, yielding 100,000 labeled latent-space vectors. %

For each attribute, we fit a linear SVM to predict the label based on the latent-space point\,---\,$\zz$ for traditional and $\ww$ for style-based\,---\,and classify the points by this plane. 
We then compute the conditional entropy ${\mathrm H}(Y|X)$ where $X$ are the classes predicted by the SVM and $Y$ are the classes determined by the pre-trained classifier. 
This tells how much additional information is required to determine the true class of a sample, given that we know on which side of the hyperplane it lies.
A low value suggests consistent latent space directions for the corresponding factor(s) of variation.

We calculate the final separability score as $\exp(\sum_i{\mathrm H}(Y_i|X_i))$, where $i$ enumerates the~40 attributes. Similar to the inception score~\cite{Salimans2016B}, the exponentiation brings the values from logarithmic to linear domain so that they are easier to compare.

Tables~\ref{tab:disentangle} and~\ref{tab:mapping} show that $\WW$ is consistently better separable than $\ZZ$, suggesting a less entangled representation. Furthermore, increasing the depth of the mapping network improves both image quality and separability in $\WW$, which is in line with the hypothesis that the synthesis network inherently favors a disentangled input representation. Interestingly, adding a mapping network in front of a traditional generator results in severe loss of separability in $\ZZ$ but improves the situation in the intermediate latent space $\WW$, and the FID improves as well. This shows that even the traditional generator architecture performs better when we introduce an intermediate latent space that does not have to follow the distribution of the training data.

\section{Conclusion}

Based on both our results and parallel work by Chen et al.~\cite{Chen2018self}, it is becoming clear that the traditional GAN generator architecture is in every way inferior to a style-based design. This is true in terms of established quality metrics, and we further believe that our investigations to the separation of high-level attributes and stochastic effects, as well as the linearity of the intermediate latent space will prove fruitful in improving the understanding and controllability of GAN synthesis. %

We note that our average path length metric could easily be used as a regularizer during training, and perhaps some variant of the linear separability metric could act as one, too. 
In general, we expect that methods for directly shaping the intermediate latent space during training will provide interesting avenues for future work.

\section{\FINAL{Acknowledgements}}
\FINAL{We thank Jaakko Lehtinen, David Luebke, and Tuomas Kynk\"a\"anniemi for in-depth discussions and helpful comments; Janne Hellsten, Tero Kuosmanen, and Pekka J\"anis for compute infrastructure and help with the code release.}

\appendix
\section{The FFHQ dataset}
\label{sec:FFHQ}
We have collected a new dataset of human faces, Flickr-Faces-HQ (FFHQ), consisting of 70,000 high-quality images at $1024^2$ resolution (Figure~\ref{fig:FFHQ}). The dataset includes vastly more variation than \textsc{CelebA-HQ} \cite{Karras2017} in terms of age, ethnicity and image background, and also has much better coverage of accessories such as eyeglasses, sunglasses, hats, etc. The images were crawled from Flickr (thus inheriting all the biases of that website) and automatically aligned \cite{Kazemi2014} and cropped. Only images under permissive licenses were collected. Various automatic filters were used to prune the set, and finally Mechanical Turk allowed us to remove the occasional statues, paintings, or photos of photos. We \FINAL{have made} the dataset publicly available at \FINAL{\ffhqlink}

\section{Truncation trick in $\WW$}
\label{sec:truncation}

If we consider the distribution of training data, it is clear that areas of low density are poorly represented and thus likely to be difficult for the generator to learn. This is a significant open problem in all generative modeling techniques.
However, it is known that drawing latent vectors from a \FINAL{truncated~\cite{Marchesi2017,Brock2018}} or otherwise shrunk~\cite{Kingma2018} sampling space tends to improve average image quality, although some amount of variation is lost.

We can follow a similar strategy. To begin, we compute the center of mass of $\WW$ as \mbox{$\barww = \EX_{\zz\sim P(\zz)}[f(\zz)]$}. In case of FFHQ this point represents a sort of an average face (Figure~\ref{fig:stylescale}, $\psi=0$). We can then scale the deviation of a given $\ww$ from the center as \mbox{$\ww' = \barww + \psi(\ww - \barww)$}, where $\psi < 1$. While Brock et al.~\cite{Brock2018} observe that only a subset of networks is amenable to such truncation even when orthogonal regularization is used, truncation in $\WW$ space seems to work reliably even without changes to the loss function.

\figstylescale

\ifarxiv
\input{supplemental-body}
\fi

{\small
\bibliographystyle{ieee}
\bibliography{paper}
}

\end{document}

%% file: figures.tex
\newcommand{\h}{0mm}
\newcommand{\hh}{0mm}
\newcommand{\hhh}{0mm}
\newcommand{\vv}{0mm}

\newcommand{\s}{\hphantom{0}}

\newcommand{\figarch}{
\begin{figure}[t]
\centering
\includegraphics[width=1.015\linewidth]{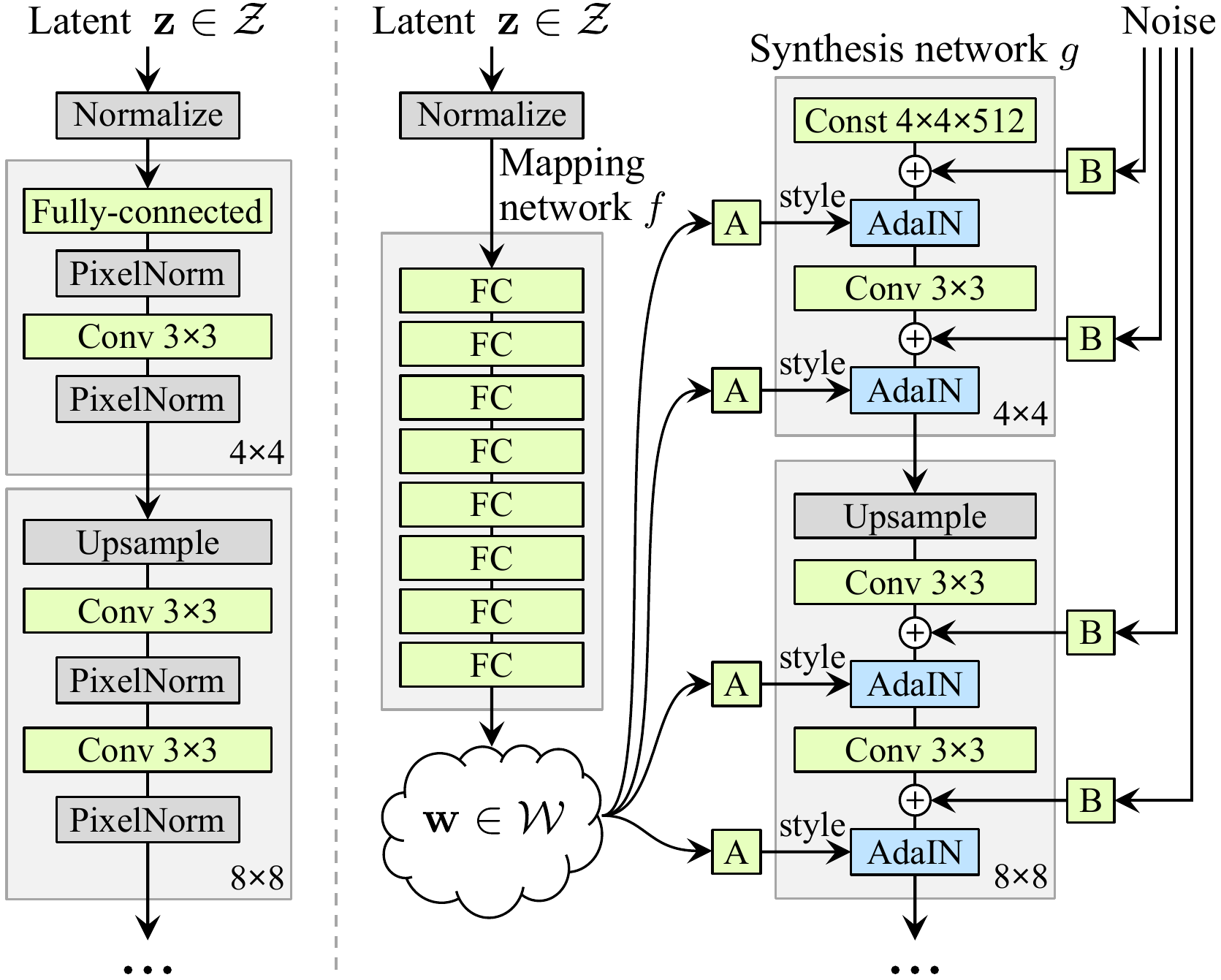}\\
\makebox[0.25\linewidth][c]{\footnotesize{(a) Traditional}}\hfill%
\makebox[0.75\linewidth][c]{\footnotesize{(b) Style-based generator}}\\
\vspace{1.9mm}
\caption{While a traditional generator \cite{Karras2017} feeds the latent code though the input layer only, we first map the input to an intermediate latent space $\WW$, which then controls the generator through adaptive instance normalization (AdaIN) at each convolution layer. Gaussian noise is added after each convolution, before evaluating the nonlinearity. Here ``A'' stands for a learned affine transform, and ``B'' applies learned per-channel scaling factors to the noise input. The mapping network $f$ consists of 8 layers and the synthesis network $g$ consists of 18 layers\,---\,two for each resolution ($4^2-1024^2$). The output of the last layer is converted to RGB using a separate $1\times1$ convolution, similar to Karras~et~al.~\cite{Karras2017}. Our generator has a total of 26.2M trainable parameters, compared to 23.1M in the traditional generator.
}
\label{fig:arch}
\vspace{-1mm}
\end{figure}
}

\newcommand{\figFFHQ}{
\begin{figure*}
\centering
\renewcommand{\h}{0.141\linewidth}
\includegraphics[width=\h]{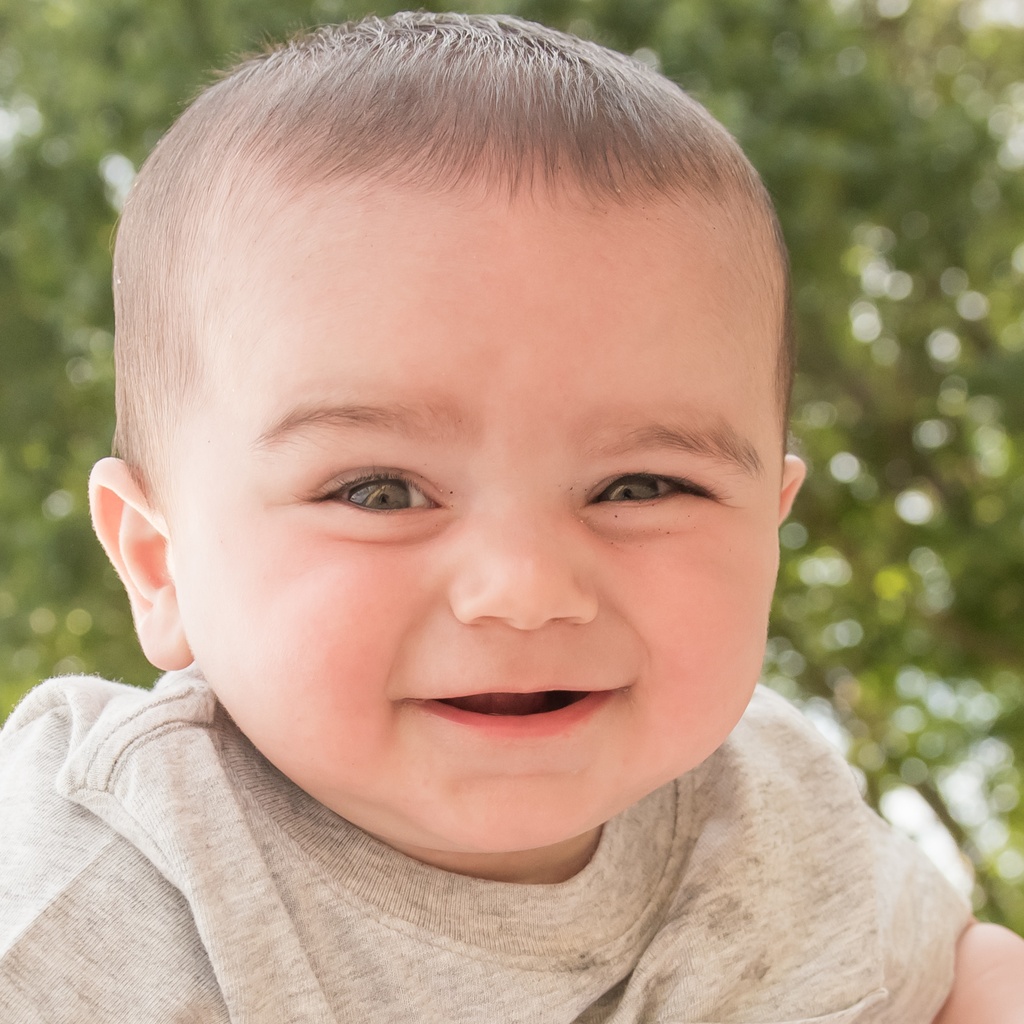}\hfill%
\includegraphics[width=\h]{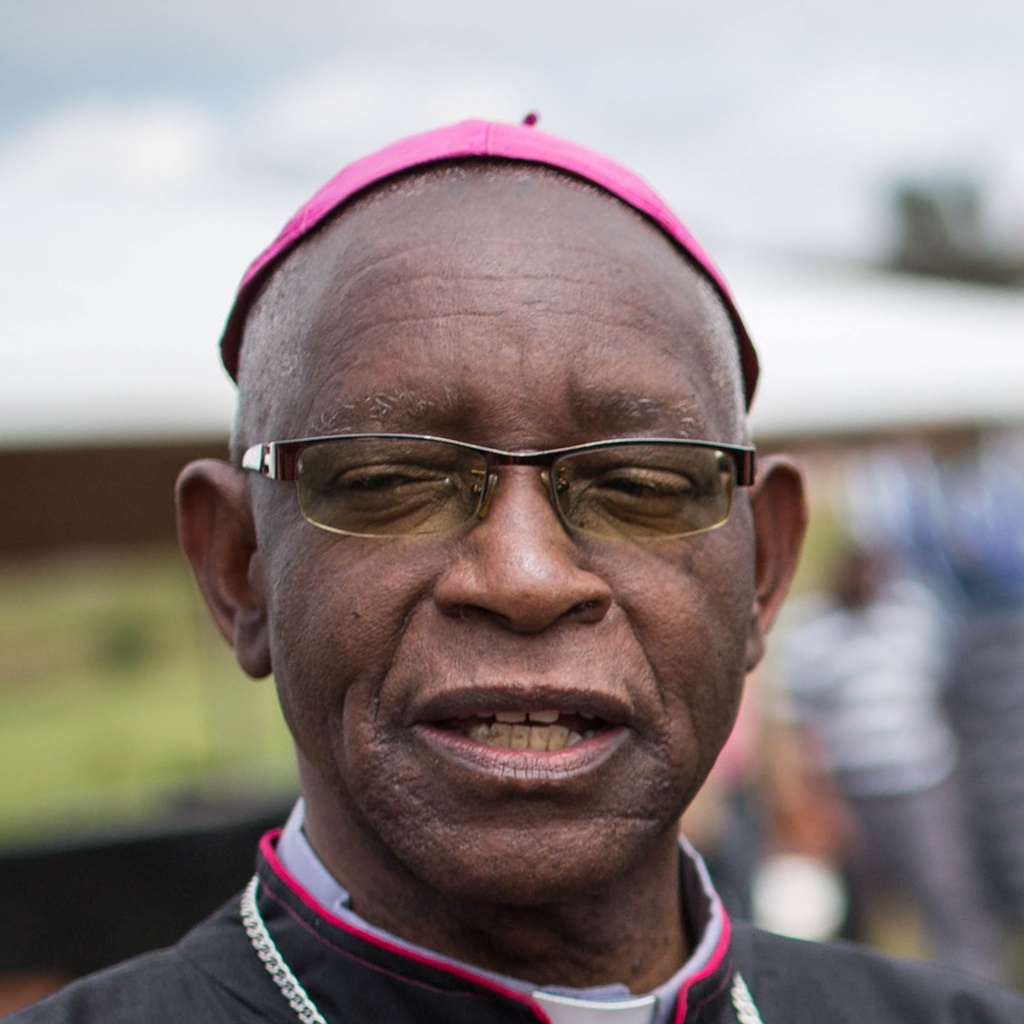}\hfill%
\includegraphics[width=\h]{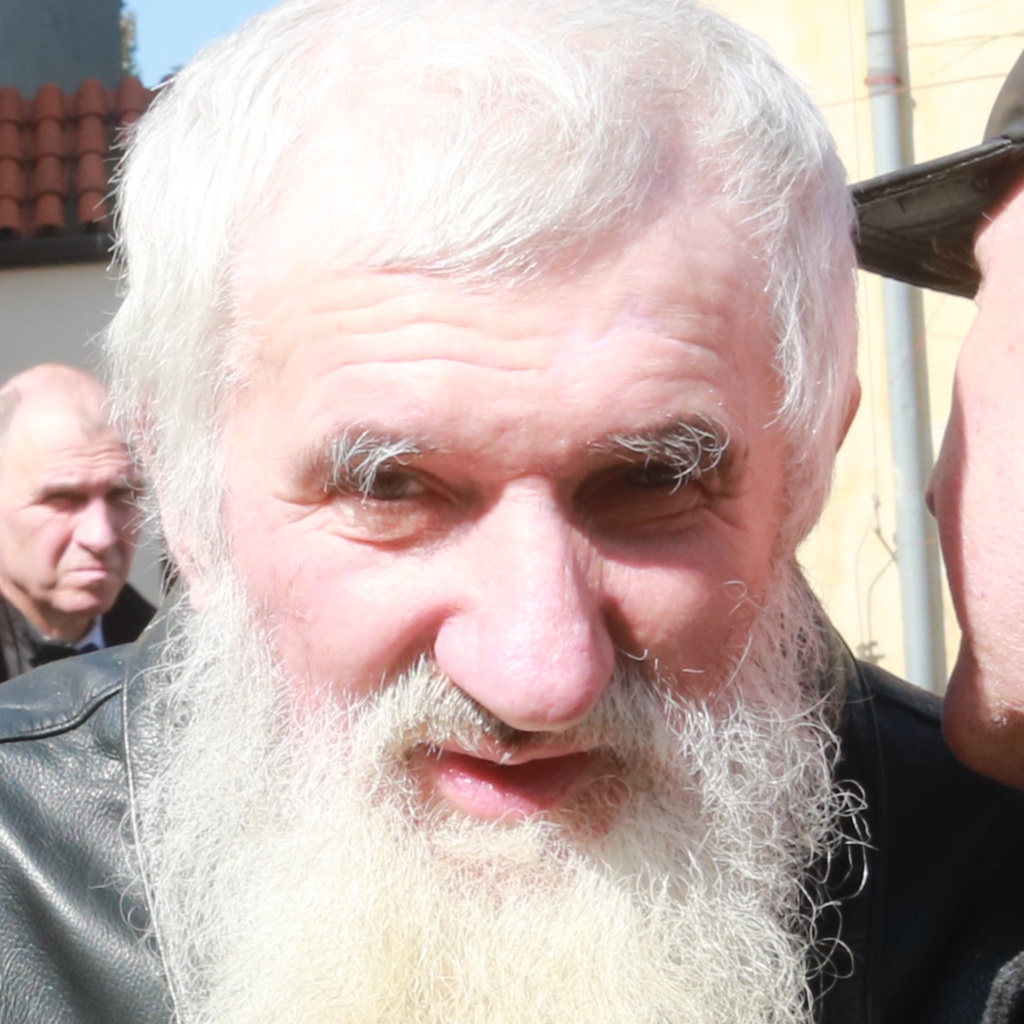}\hfill%
\includegraphics[width=\h]{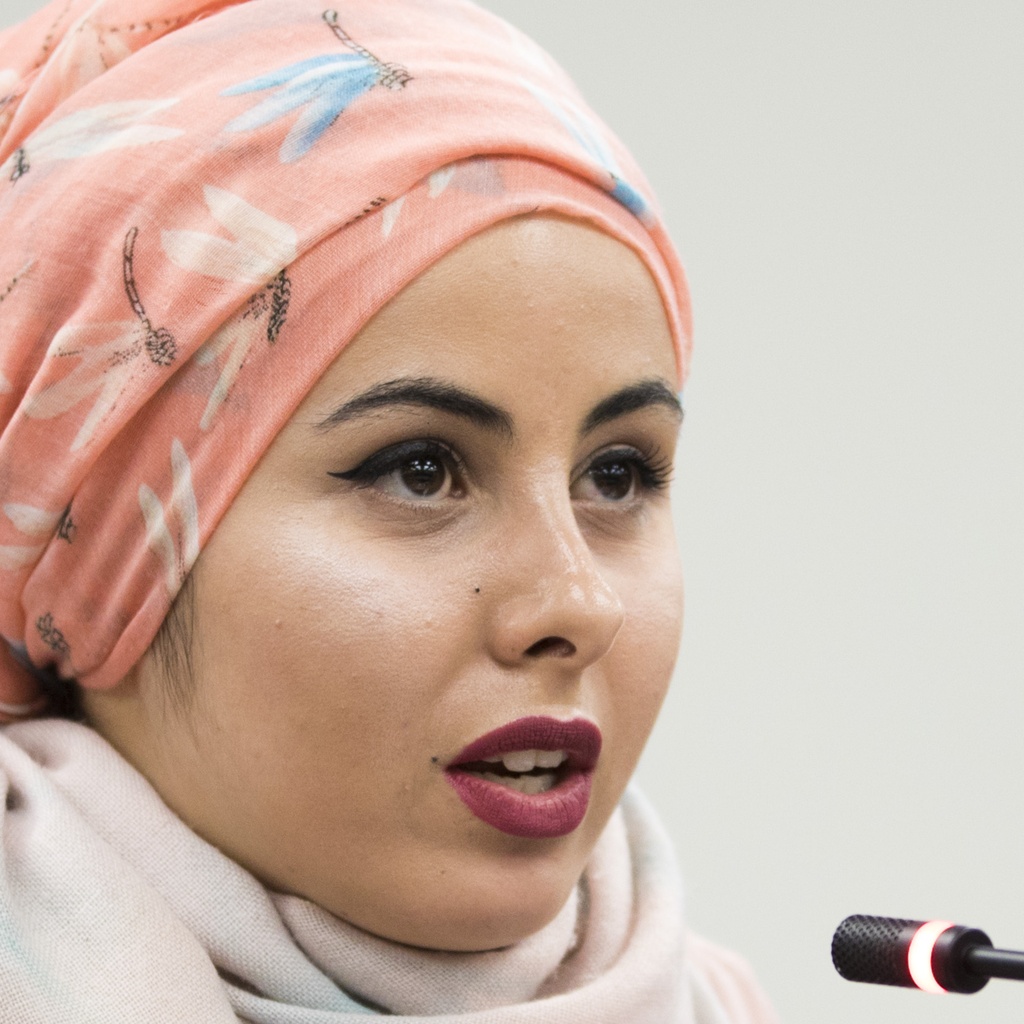}\hfill%
\includegraphics[width=\h]{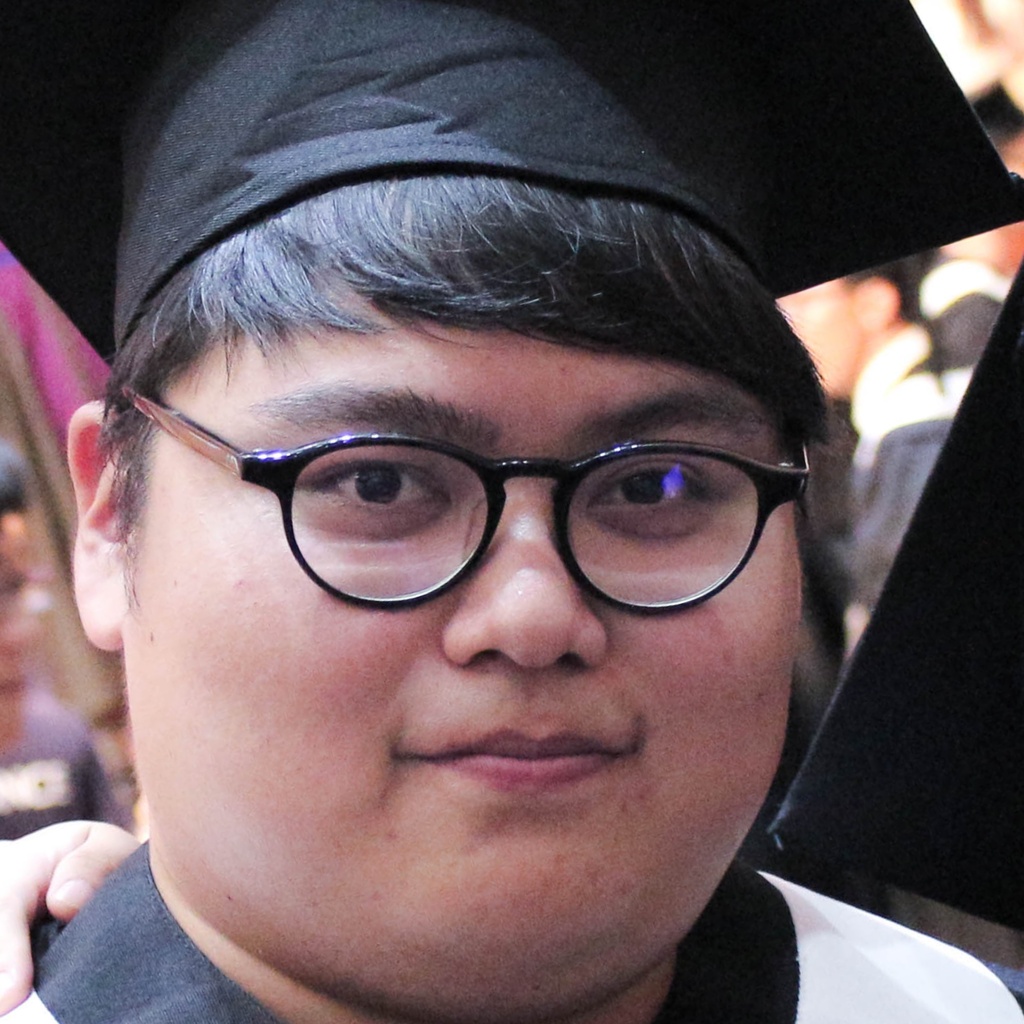}\hfill%
\includegraphics[width=\h]{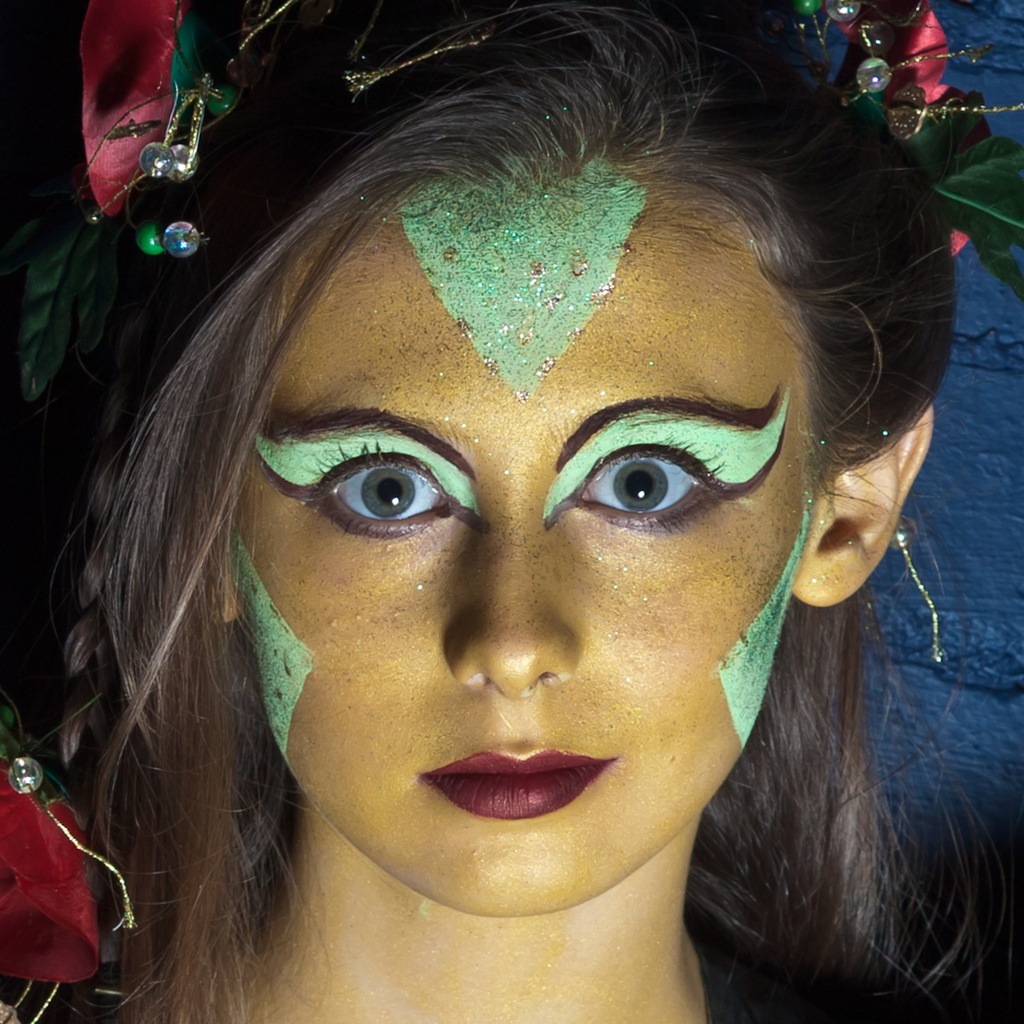}\hfill%
\includegraphics[width=\h]{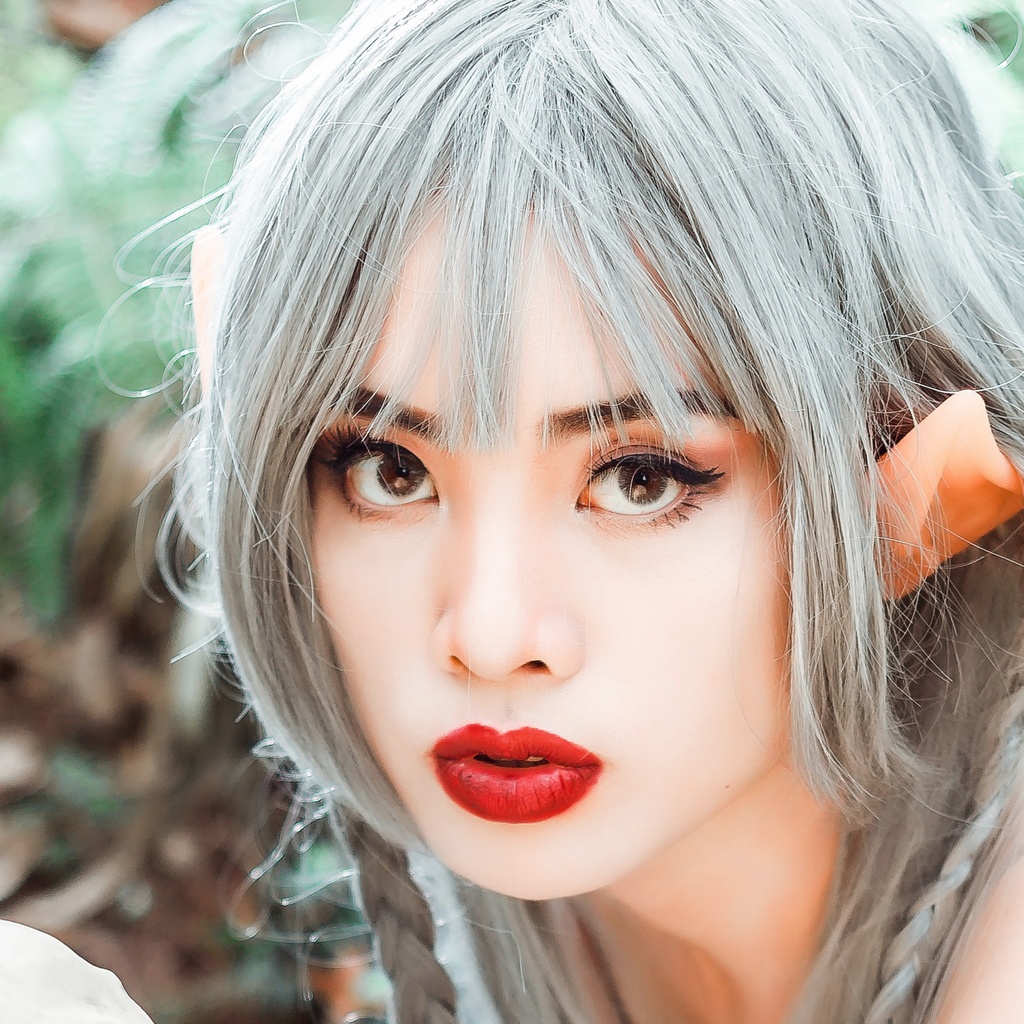}\hfill%
\caption{The FFHQ dataset offers a lot of variety in terms of age, ethnicity, viewpoint, lighting, and image background.%
}
\label{fig:FFHQ}
\vspace{-1mm}
\end{figure*}
}

\newcommand{\tabFID}{
\begin{table}[t]
\centering
\newcolumntype{x}{>{\centering\arraybackslash\hspace{0pt}}p{15mm}}
\footnotesize{
\begin{tabular}{|l@{\hspace{1.5mm}}l|x|x|}
\hline
            & \textbf{Method}                               & \textbf{CelebA-HQ}    & \textbf{FFHQ} \\\hline\hline
\arch{a}    & Baseline Progressive GAN \cite{Karras2017}\ \ & 7.79                  & 8.04          \\\hline %
\arch{b}    & + Tuning (incl. bilinear up/down)         & 6.11                  & 5.25          \\\hline %
\arch{c}    & + Add mapping and styles                      & 5.34                  & 4.85          \\\hline %
\arch{d}    & + Remove traditional input                    & 5.07                  & 4.88          \\\hline %
\arch{e}    & + Add noise inputs                            & \textbf{5.06}         & 4.42          \\\hline %
\arch{f}    & + Mixing regularization                       & 5.17                  & \textbf{4.40} \\\hline %
\end{tabular}
}\vspace{1.5mm}
\caption{Fr\'echet inception distance (FID) for various generator designs (lower is better). In this paper we calculate the FIDs using 50,000 images drawn randomly from the training set, and report the lowest distance encountered over the course of training. 
}
\label{tab:FID}
\vspace{-1.4mm}
\end{table}
}

\newcommand{\tabmixing}{
\begin{table}[t]
\centering
\newcolumntype{x}{>{\centering\arraybackslash\hspace{0pt}}p{10.2mm}}
\footnotesize{
\begin{tabular}{|l@{\hspace{1.5mm}}l|x|x|x|x|}
\hline
            & \textbf{Mixing}         & \multicolumn{4}{c|}{\textbf{Number of latents during testing}}          \\
            & \textbf{regularization} & \textbf{1}      & \textbf{2}    & \textbf{3}        & \textbf{4}        \\\hline\hline
\arch{e}    & 0\%                     & 4.42            & 8.22          & 12.88             & 17.41             \\\hline %
            & 50\%                    & 4.41            & 6.10          & \s8.71            & 11.61             \\\hline %
\arch{f}    & 90\%                    & \textbf{4.40}   & \textbf{5.11} & \s6.88            & \s9.03            \\\hline %
            & 100\%                   & 4.83            & 5.17          & \textbf{\s6.63}   & \textbf{\s8.40}   \\\hline %
\end{tabular}
}\vspace{1.5mm}\extraspaceafter
\caption{FIDs in FFHQ for networks trained by enabling the mixing regularization for different percentage of training examples. Here we stress test the trained networks by randomizing $1\ldots4$ latents and the crossover points between them. Mixing regularization improves the tolerance to these adverse operations significantly. Labels \arch{e} and \arch{f} refer to the configurations in Table~\protect\ref{tab:FID}.
}
\label{tab:mixing}
\end{table}
}

\newcommand{\tabdisentangle}{
\begin{table}[t]
\centering
\newcolumntype{x}{>{\centering\arraybackslash\hspace{0pt}}p{11.2mm}}
\footnotesize{
\begin{tabular}{|c@{\hspace{1.5mm}}l@{\hspace{1mm}}c|x|x|x|}
\hline
            & \multirow{2}{*}{\textbf{Method}}  &           & \multicolumn{2}{c|}{\textbf{Path length}} & \textbf{Separa-}\\
            &                                   &           & \textbf{full}     & \textbf{end}      & \textbf{bility}   \\\hline\hline
\arch{b}    & Traditional generator             & $\ZZ$     & 412.0             & 415.3             & 10.78             \\\hline %
\arch{d}    & Style-based generator             & $\WW$     & 446.2             & 376.6             & \s3.61            \\\hline %
\arch{e}    & + Add noise inputs                & $\WW$     & \textbf{200.5}    & \textbf{160.6}    & \s3.54            \\\hline %
            & + Mixing 50\%                     & $\WW$     & 231.5             & 182.1             & \textbf{\s3.51}   \\\hline %
\arch{f}    & + Mixing 90\%                     & $\WW$     & 234.0             & 195.9             & \s3.79            \\\hline %
\end{tabular}
}\vspace{1.5mm}\extraspaceafter
\caption{Perceptual path lengths and separability scores for various generator architectures in FFHQ (lower is better). We perform the measurements in $\ZZ$ for the traditional network, and in $\WW$ for style-based ones.
Making the network resistant to style mixing appears to distort the intermediate latent space $\WW$ somewhat. 
We hypothesize that mixing makes it more difficult for $\WW$ to efficiently encode factors of variation that span multiple scales.
}
\label{tab:disentangle}
\vspace*{1mm}
\end{table}
}

\newcommand{\tabmapping}{
\begin{table}[t]
\centering
\newcolumntype{x}{>{\centering\arraybackslash\hspace{0pt}}p{9.6mm}}
\footnotesize{
\begin{tabular}{|c@{\hspace{1.5mm}}l@{\hspace{1mm}}c|x|x|x|x|}
\hline
            & \multirow{2}{*}{\textbf{Method}} & & \multirow{2}{*}{\textbf{FID}}  & \multicolumn{2}{c|}{\textbf{Path length}} & \textbf{Separa-}\\
            &                   &           &               & \textbf{full}     & \textbf{end}      & \textbf{bility}   \\\hline\hline
\arch{b}    & Traditional 0     & $\ZZ$     & 5.25          & 412.0             & 415.3             & \s10.78           \\\hline %
            & Traditional 8     & $\ZZ$     & 4.87          & 896.2             & 902.0             & 170.29            \\\hline %
            & Traditional 8     & $\WW$     & 4.87          & 324.5             & 212.2             & \s\s6.52          \\\hline %
            & Style-based 0     & $\ZZ$     & 5.06          & 283.5             & 285.5             & \s\s9.88          \\\hline %
            & Style-based 1     & $\WW$     & 4.60          & 219.9             & 209.4             & \s\s6.81          \\\hline %
            & Style-based 2     & $\WW$     & 4.43          & \textbf{217.8}    & 199.9             & \s\s6.25          \\\hline %
\arch{f}    & Style-based 8     & $\WW$     & \textbf{4.40} & 234.0             & \textbf{195.9}    & \textbf{\s\s3.79} \\\hline %
\end{tabular}
}\vspace{1.5mm}\extraspaceafter
\caption{\label{tab:mapping}%
The effect of a mapping network in FFHQ. The number in method name indicates the depth of the mapping network. We see that FID, separability, and path length all benefit from having a mapping network, and this holds for both style-based and traditional generator architectures. Furthermore, a deeper mapping network generally performs better than a shallow one.
}
\end{table}
}

\newcommand{\figpathlenbehavior}{
\begin{figure}[t]
\centering
\includegraphics[width=\linewidth]{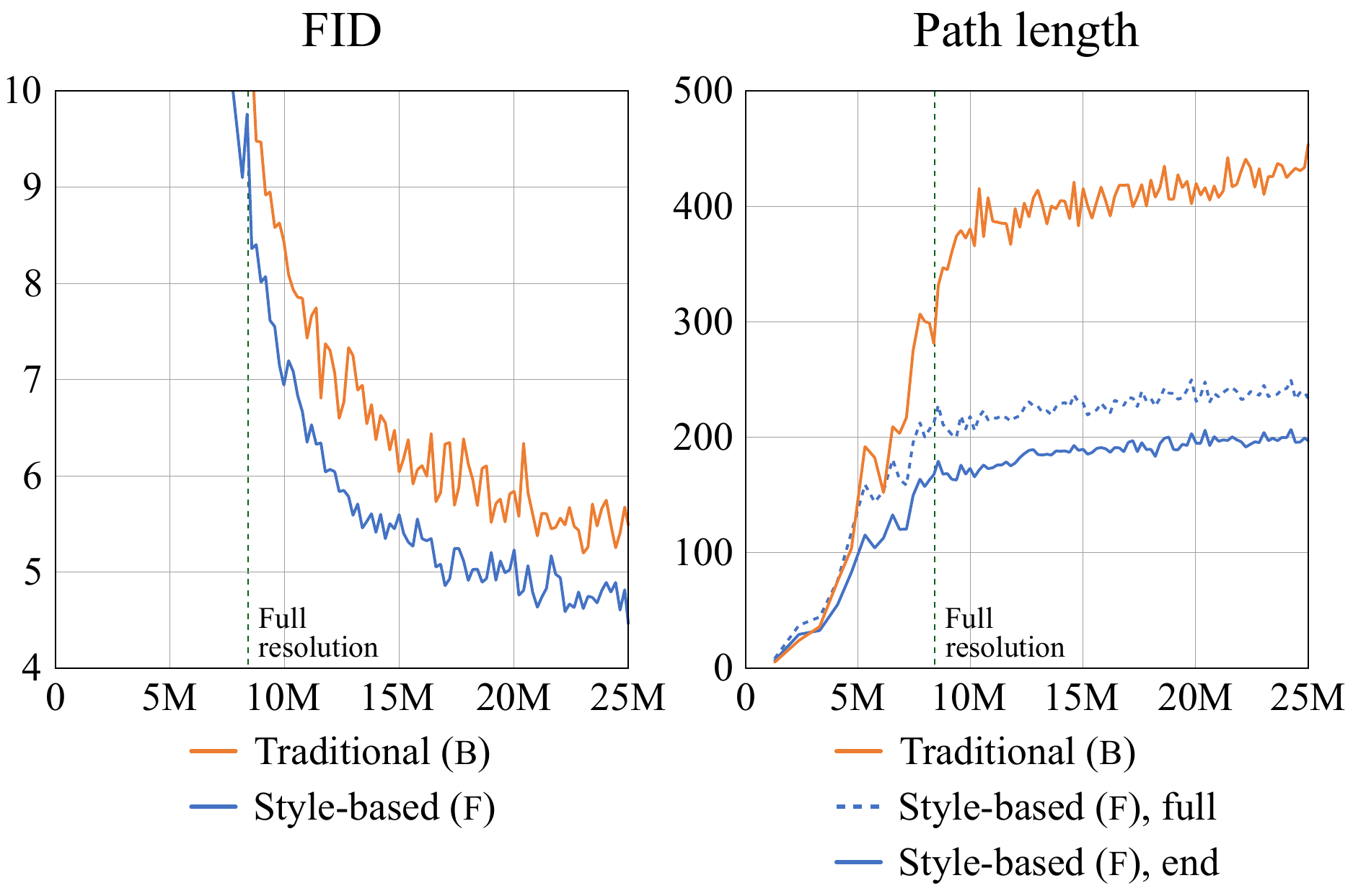}%
\caption{\label{fig:pathlenbehavior}%
FID and perceptual path length metrics over the course of training in our configurations \arch{b} and \arch{f} using the FFHQ dataset.
Horizontal axis denotes the number of training images seen by the discriminator.
The dashed vertical line at 8.4M images marks the point when training has progressed to full $1024^2$ resolution.
On the right, we show only one curve for the traditional generator's path length measurements, because there is no discernible difference between full-path and endpoint sampling in $\ZZ$.
}
\end{figure}
}

\newcommand{\figstylescale}{
\renewcommand{\h}{0.166\linewidth}
\renewcommand{\hh}{\hspace{0.0mm}}
\begin{figure}[t]
\raisebox{15.5mm}{\makebox[0pt][l]{%
\includegraphics[width=\h]{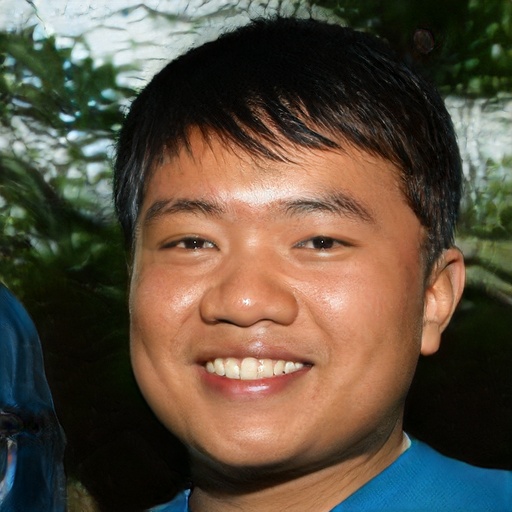}\hh%
\includegraphics[width=\h]{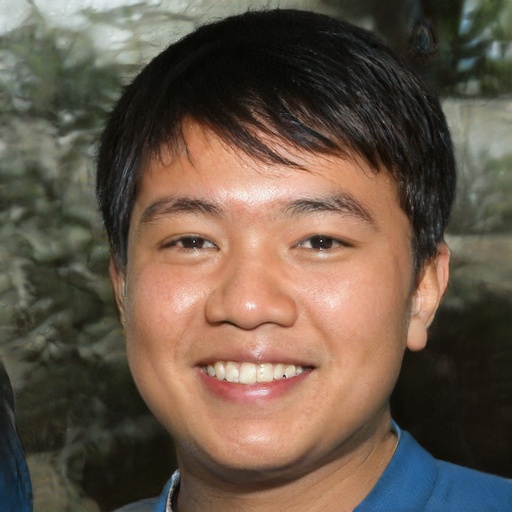}\hh%
\includegraphics[width=\h]{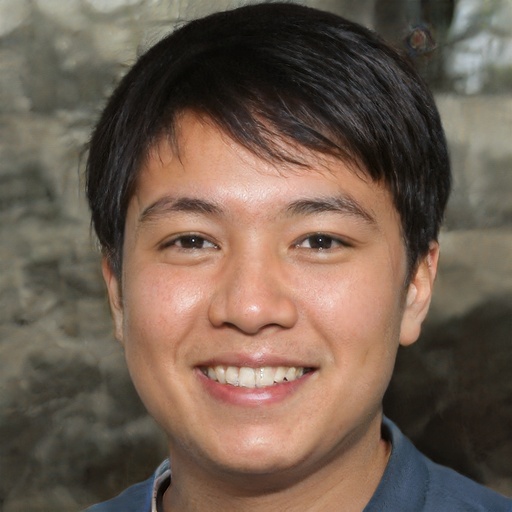}\hh%
\includegraphics[width=\h]{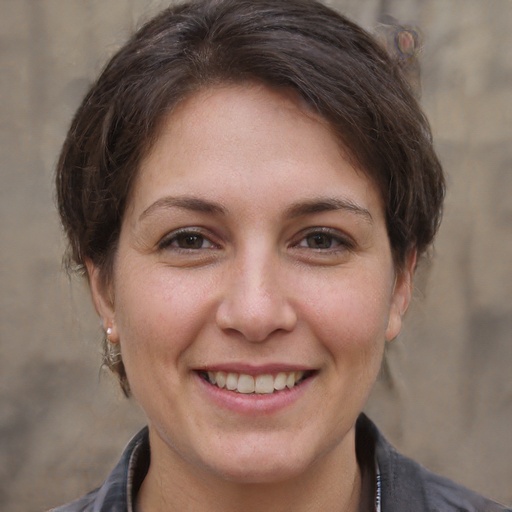}\hh%
\includegraphics[width=\h]{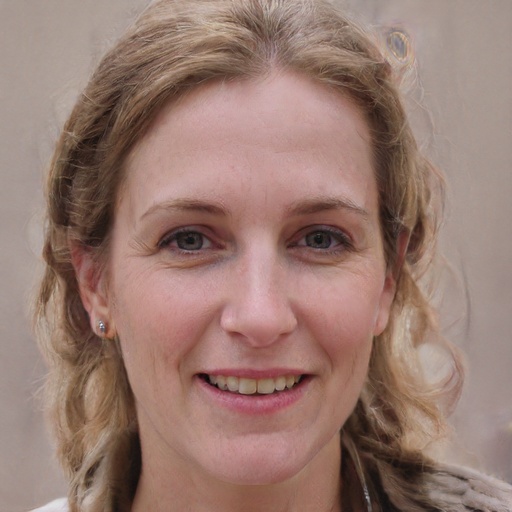}\hh%
\includegraphics[width=\h]{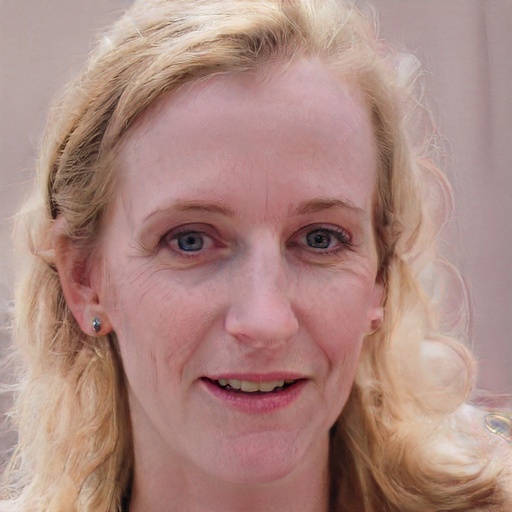}}}%
\raisebox{0mm}{\makebox{%
\includegraphics[width=\h]{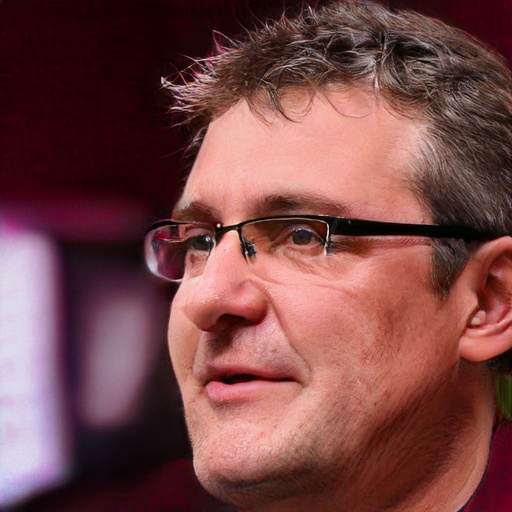}\hh%
\includegraphics[width=\h]{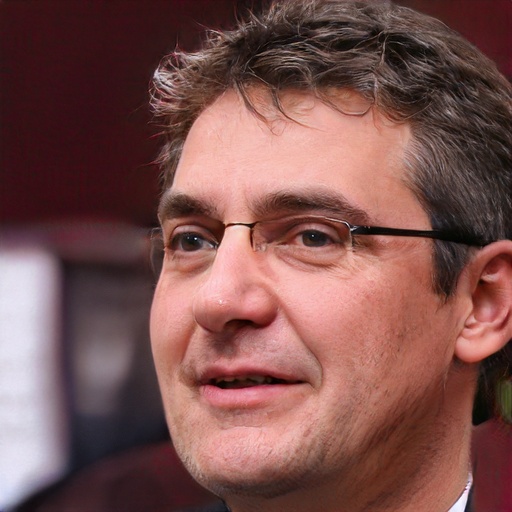}\hh%
\includegraphics[width=\h]{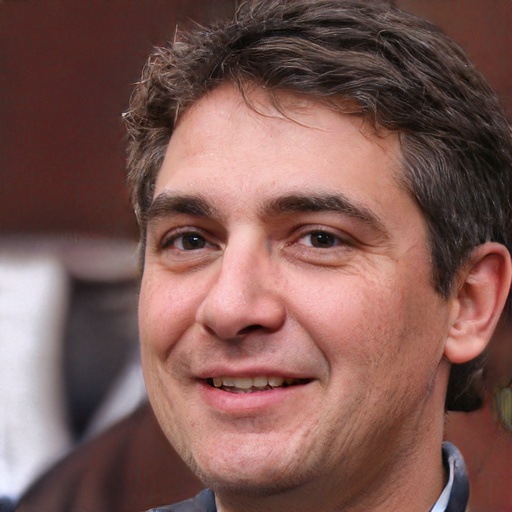}\hh%
\includegraphics[width=\h]{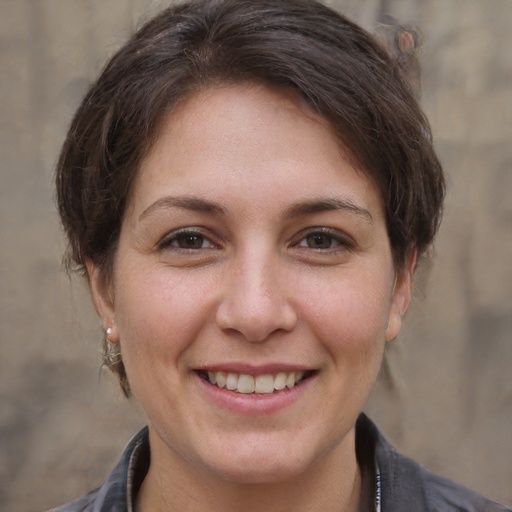}\hh%
\includegraphics[width=\h]{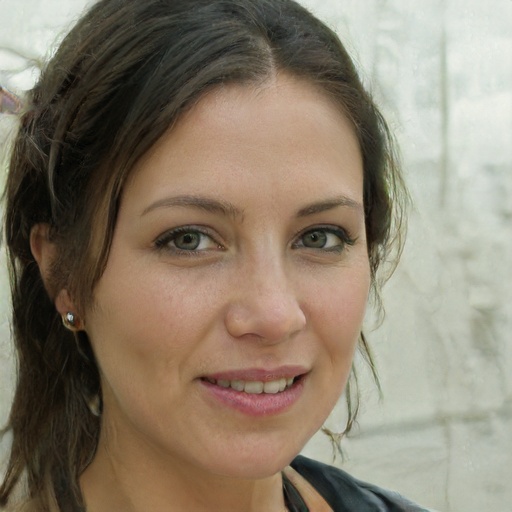}\hh%
\includegraphics[width=\h]{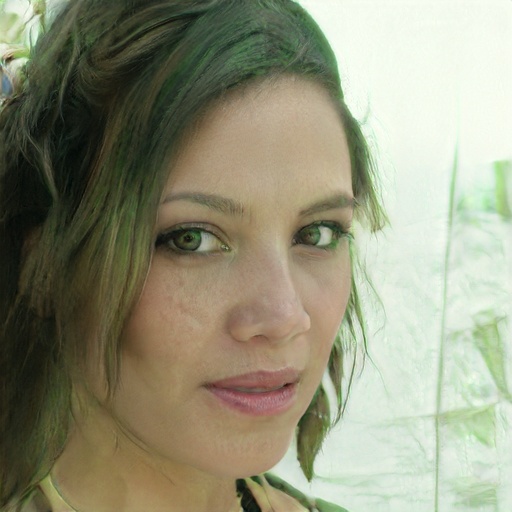}}}\\
\scriptsize{%
\makebox[\h][c]{$\psi = 1$}\hh%
\makebox[\h][c]{$\psi = 0.7$}\hh%
\makebox[\h][c]{$\psi = 0.5$}\hh%
\makebox[\h][c]{$\psi = 0$}\hh%
\makebox[\h][c]{$\psi = -0.5$}\hh%
\makebox[\h][c]{$\psi = -1$}\\
}
\caption{The effect of truncation trick as a function of style scale $\psi$. When we fade $\psi \rightarrow 0$, all faces converge to the ``mean'' face of FFHQ. This face is similar for all trained networks, and the interpolation towards it never seems to cause artifacts. By applying negative scaling to styles, we get the corresponding opposite or ``anti-face''. It is interesting that various high-level attributes often flip between the opposites, including viewpoint, glasses, age, coloring, hair length, and often gender.
}
\label{fig:stylescale}
\end{figure}
}

\newcommand{\figstylemix}{
\renewcommand{\h}{27.4mm}
\renewcommand{\hh}{2.5mm}
\renewcommand{\vv}{\vspace*{-0.35mm}}
\begin{figure*}[t]
\centering
\begin{minipage}[c]{0.972\linewidth}
\makebox[\hh][c]{}\hfill%
\makebox[\h][c]{\textbf{\normalsize{\ Source A}}}\hfill%
\rotatebox[origin=l]{90}{\makebox[0mm][l]{\hspace*{0.05\linewidth}\textbf{\normalsize{\raisebox{.5mm}[0mm][0mm]{Source B}}}}}\hfill%
\newcommand{\varA}{var639}\includegraphics[height=\h]{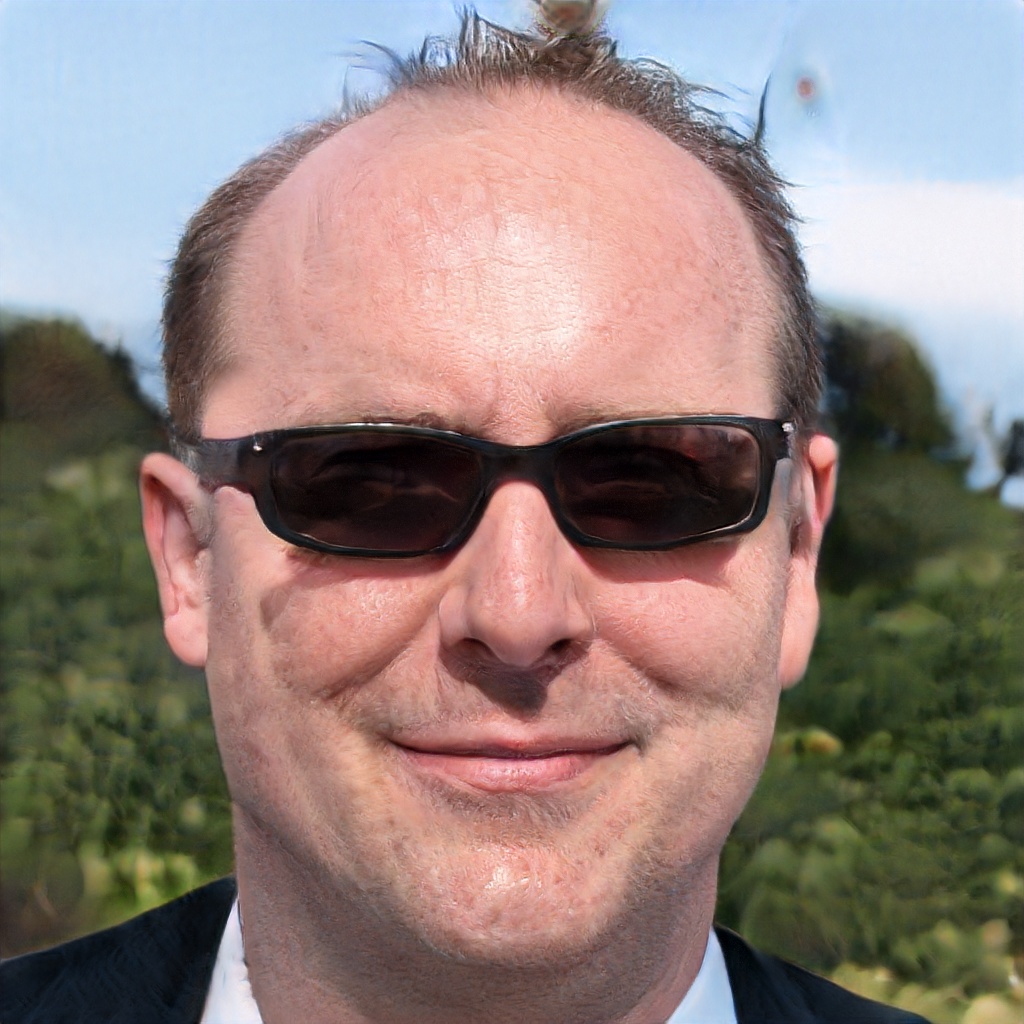}%
\newcommand{\varB}{var701}\includegraphics[height=\h]{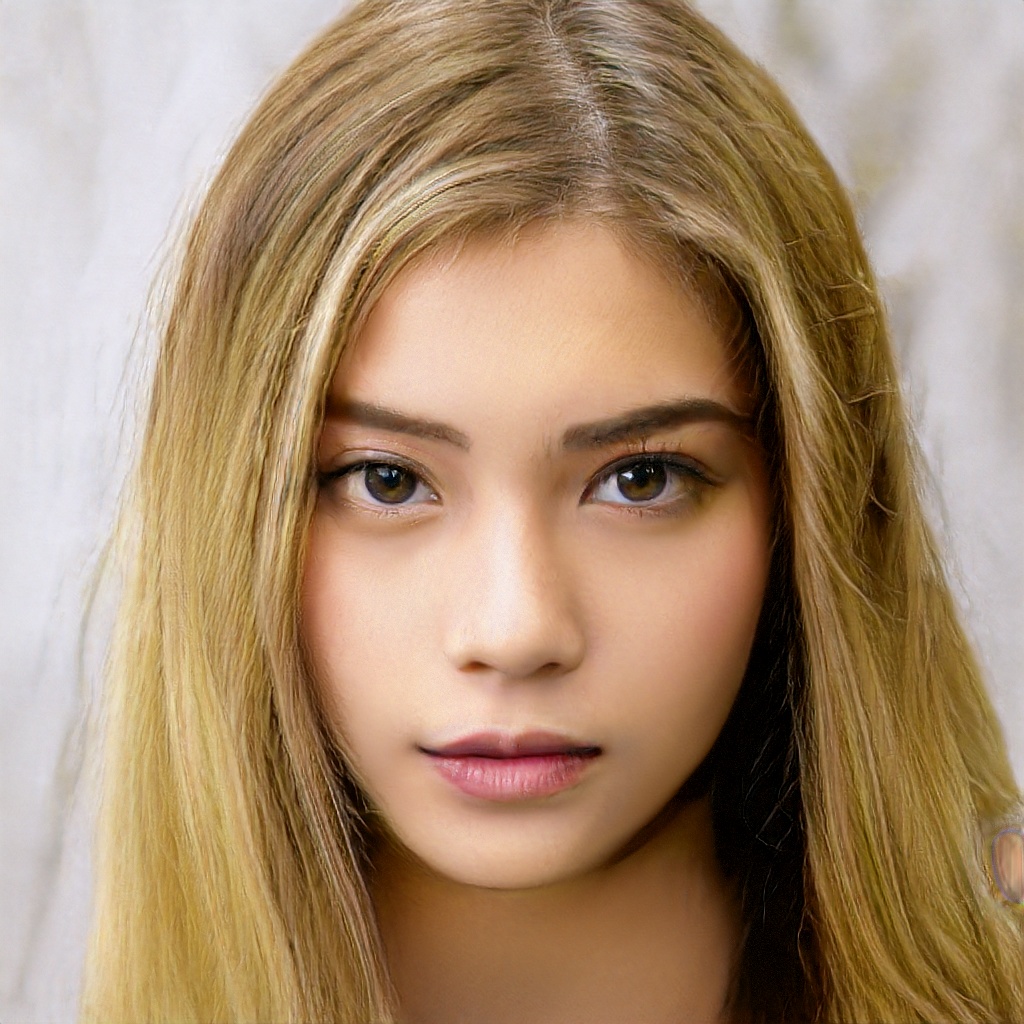}%
\newcommand{\varC}{var687}\includegraphics[height=\h]{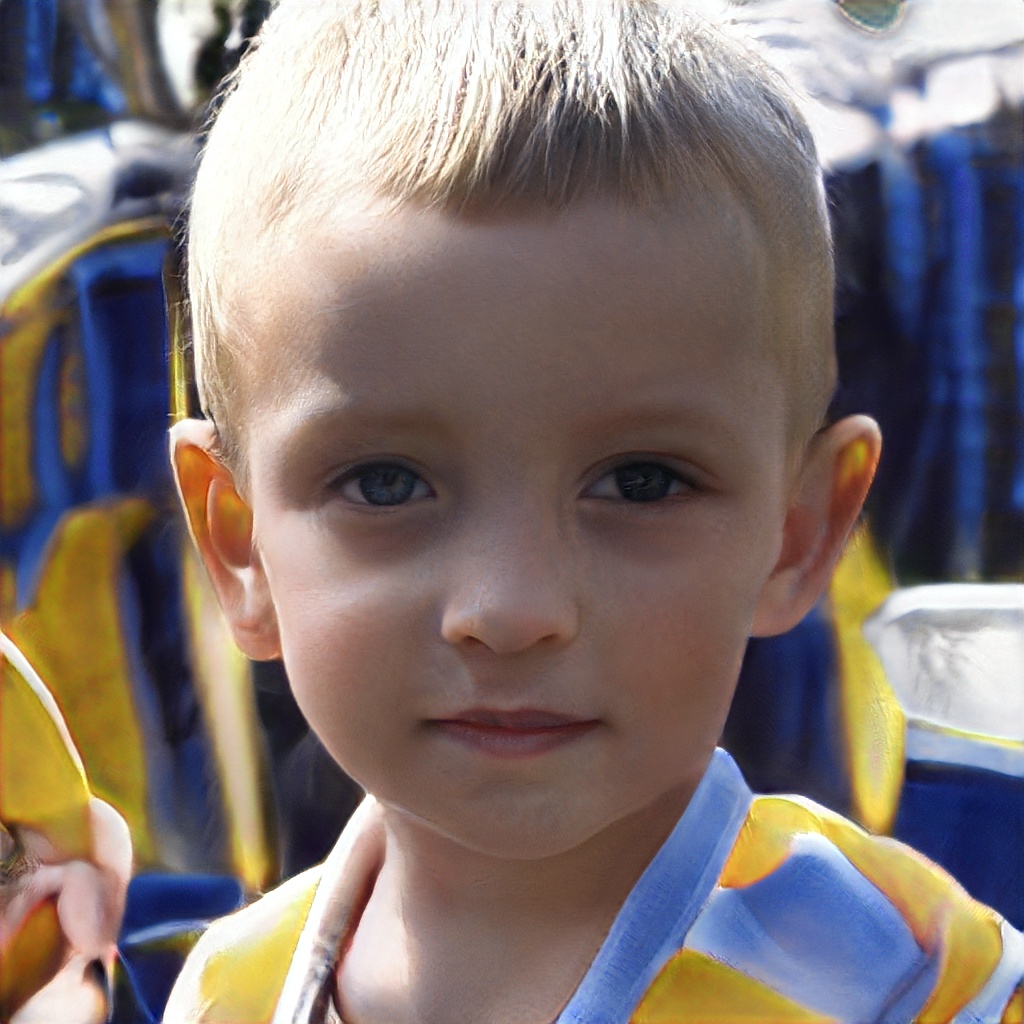}%
\newcommand{\varD}{var615}\includegraphics[height=\h]{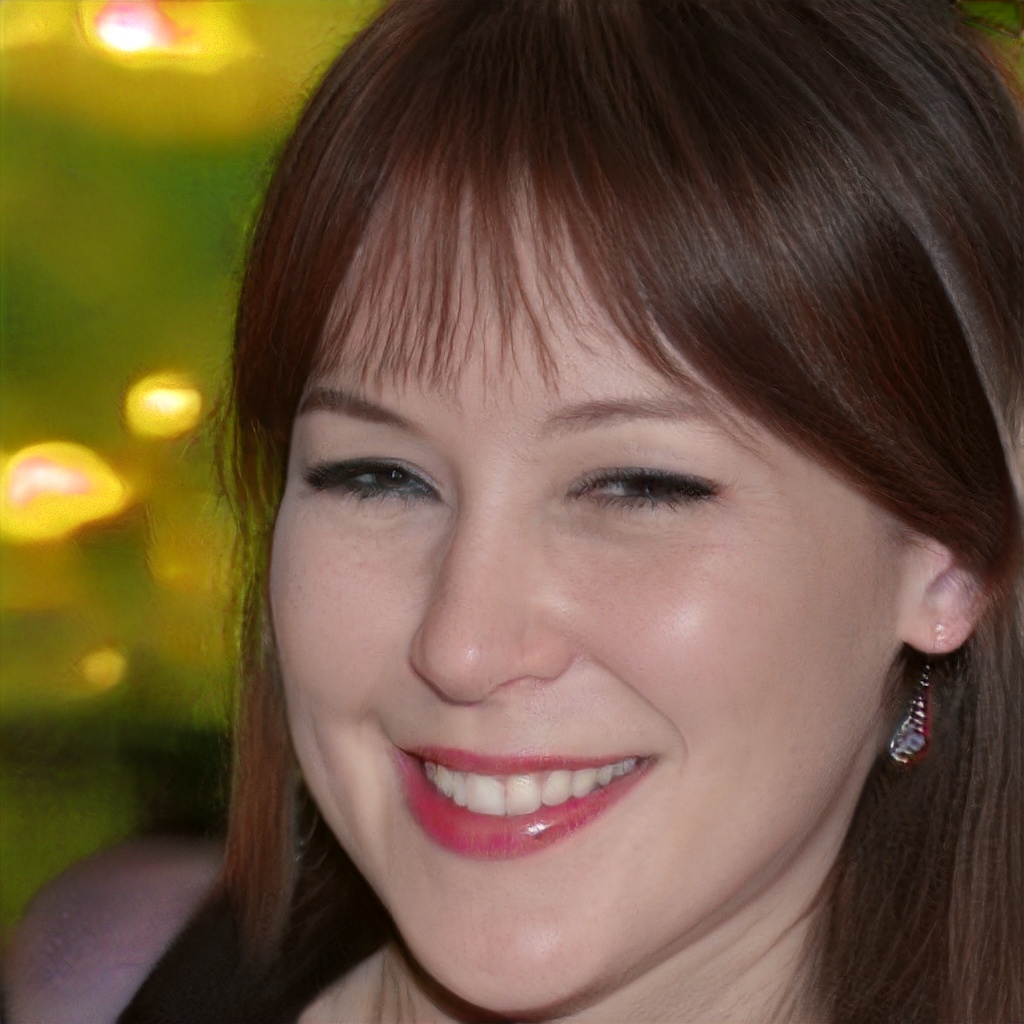}%
\newcommand{\varE}{var2268}\includegraphics[height=\h]{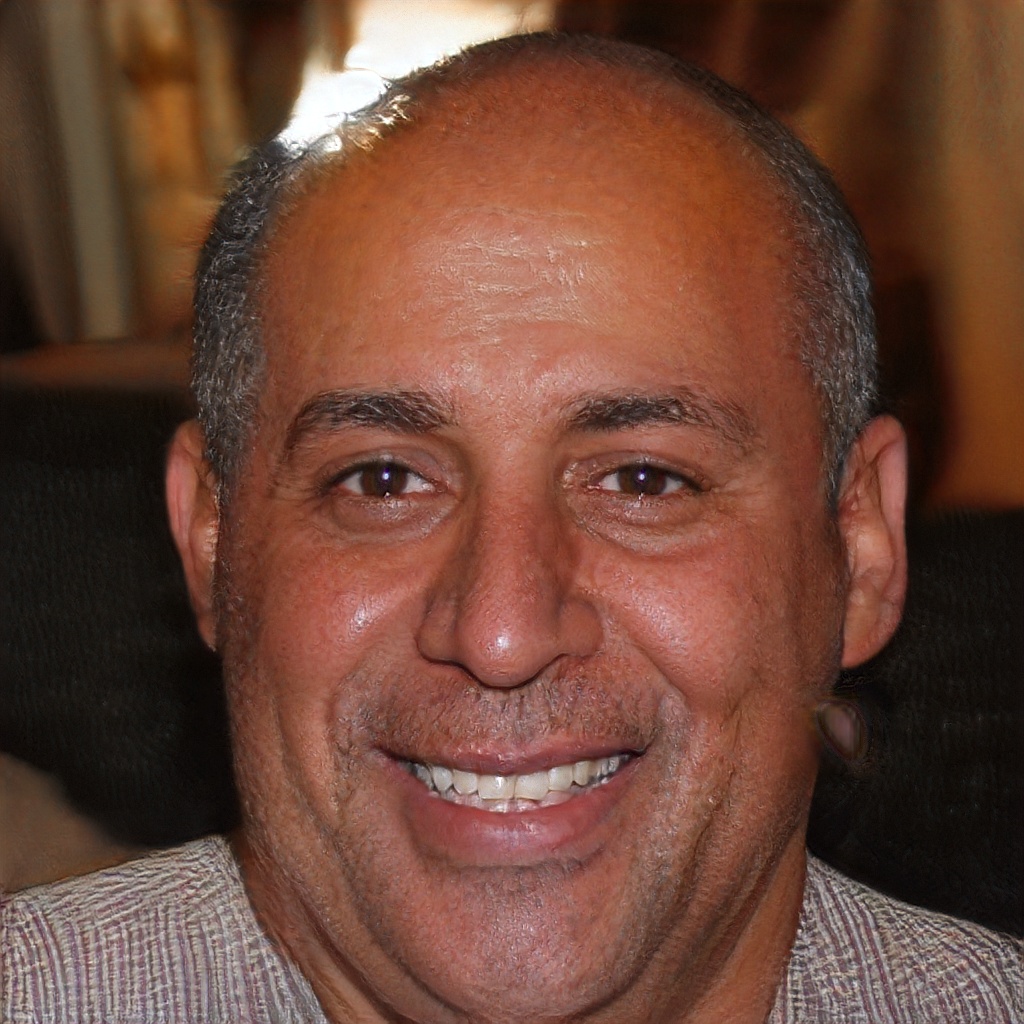}\vspace*{-3.5mm}\\
\begin{tikzpicture}\draw (0,0) -- (\linewidth,0);\end{tikzpicture}\vspace*{-0.5mm}\\%
\makebox[\hh]{\rotatebox[origin=l]{90}{\makebox[\h][c]{\hspace{-0.332\linewidth}\normalsize{Coarse styles from source B}}}}\hspace{0.5mm}%
\newcommand{\seed}{seed888}
\includegraphics[height=\h]{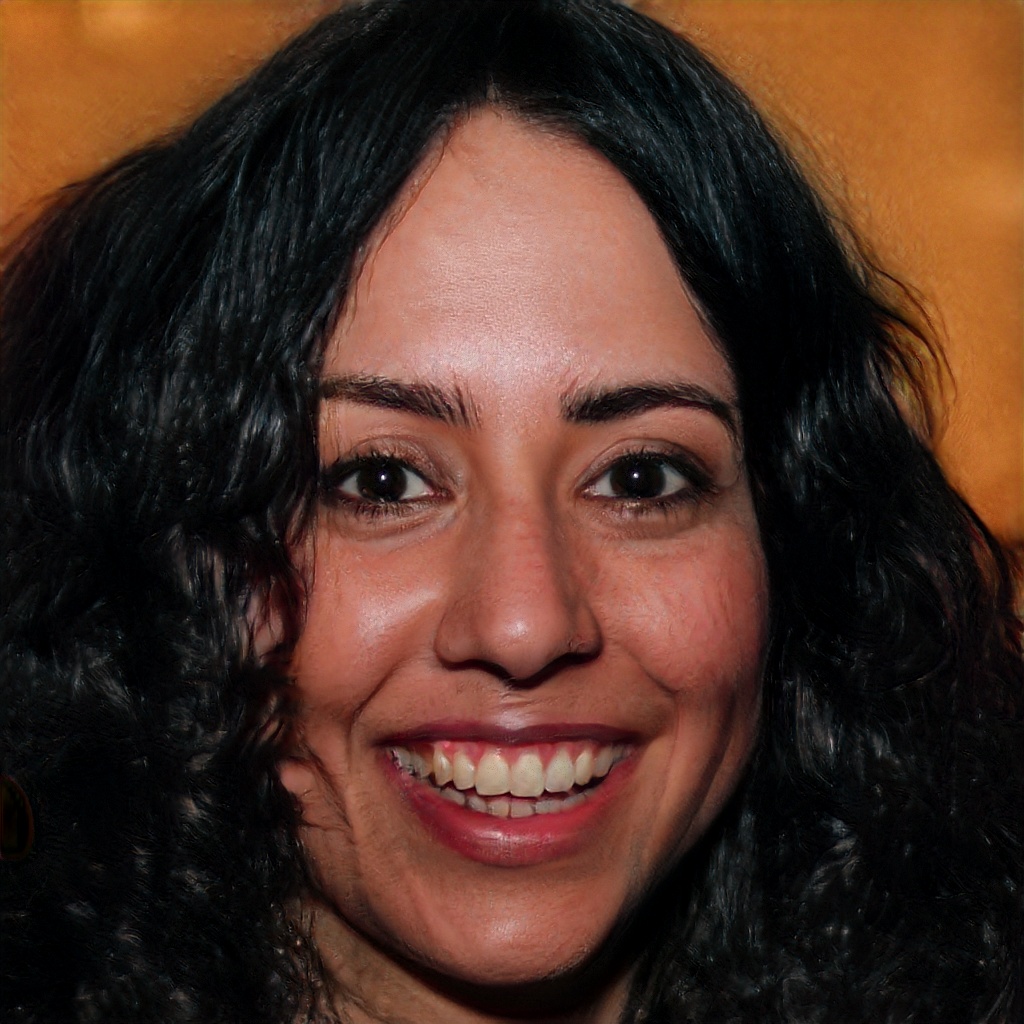}\hfill%
\includegraphics[height=\h]{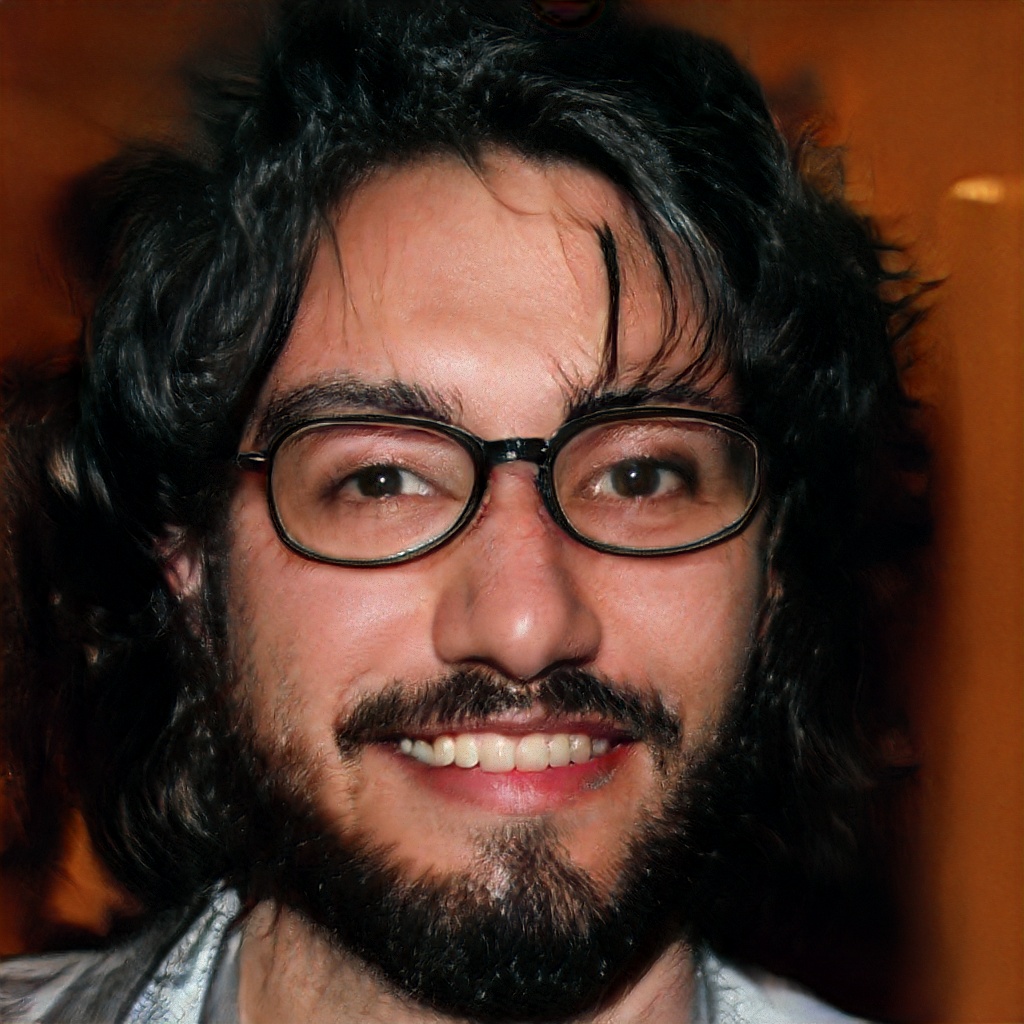}%
\includegraphics[height=\h]{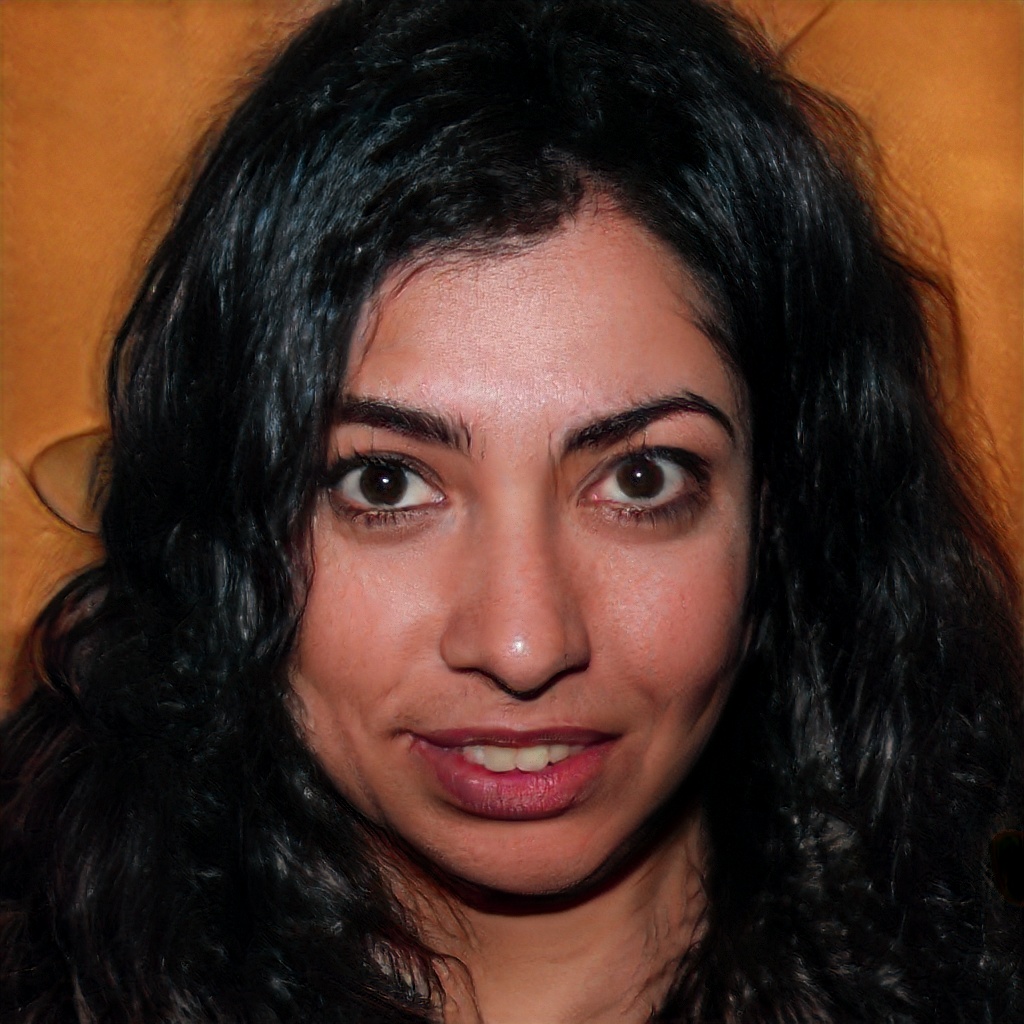}%
\includegraphics[height=\h]{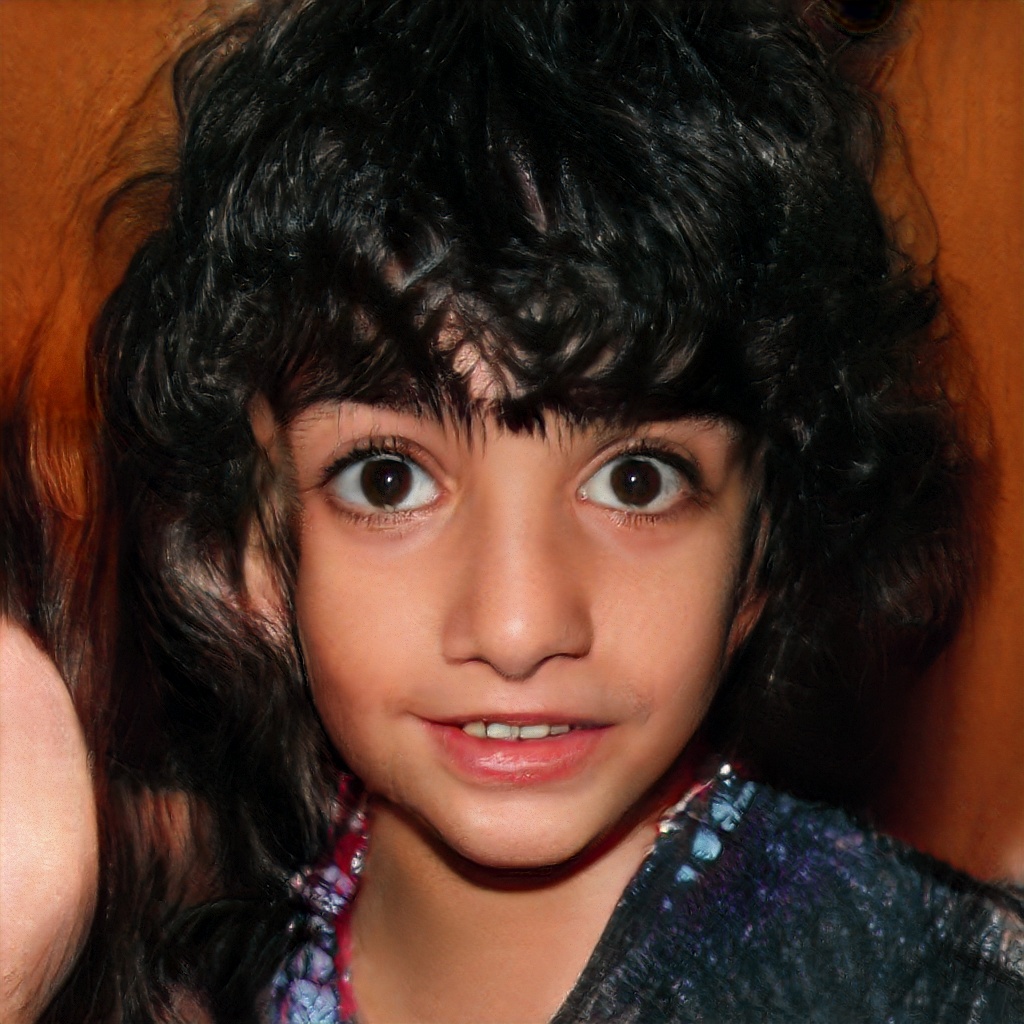}%
\includegraphics[height=\h]{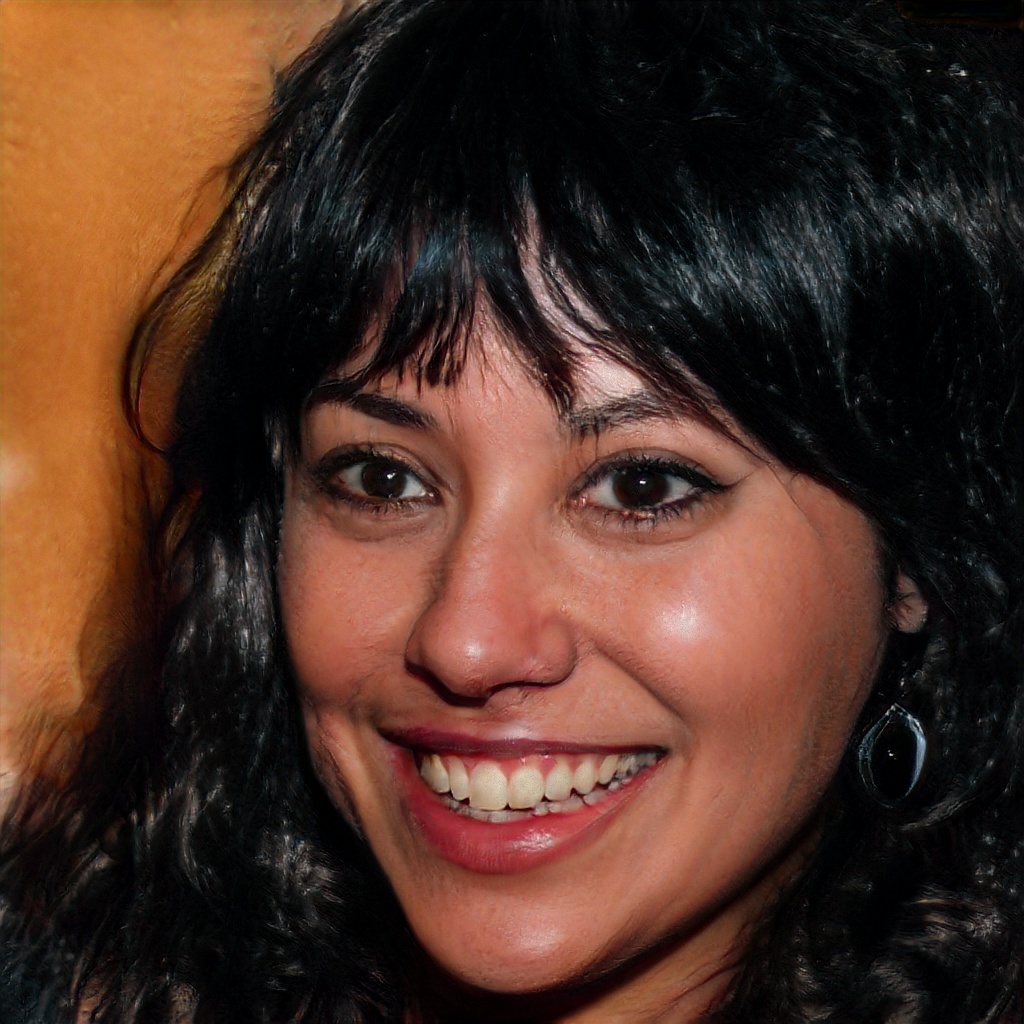}%
\includegraphics[height=\h]{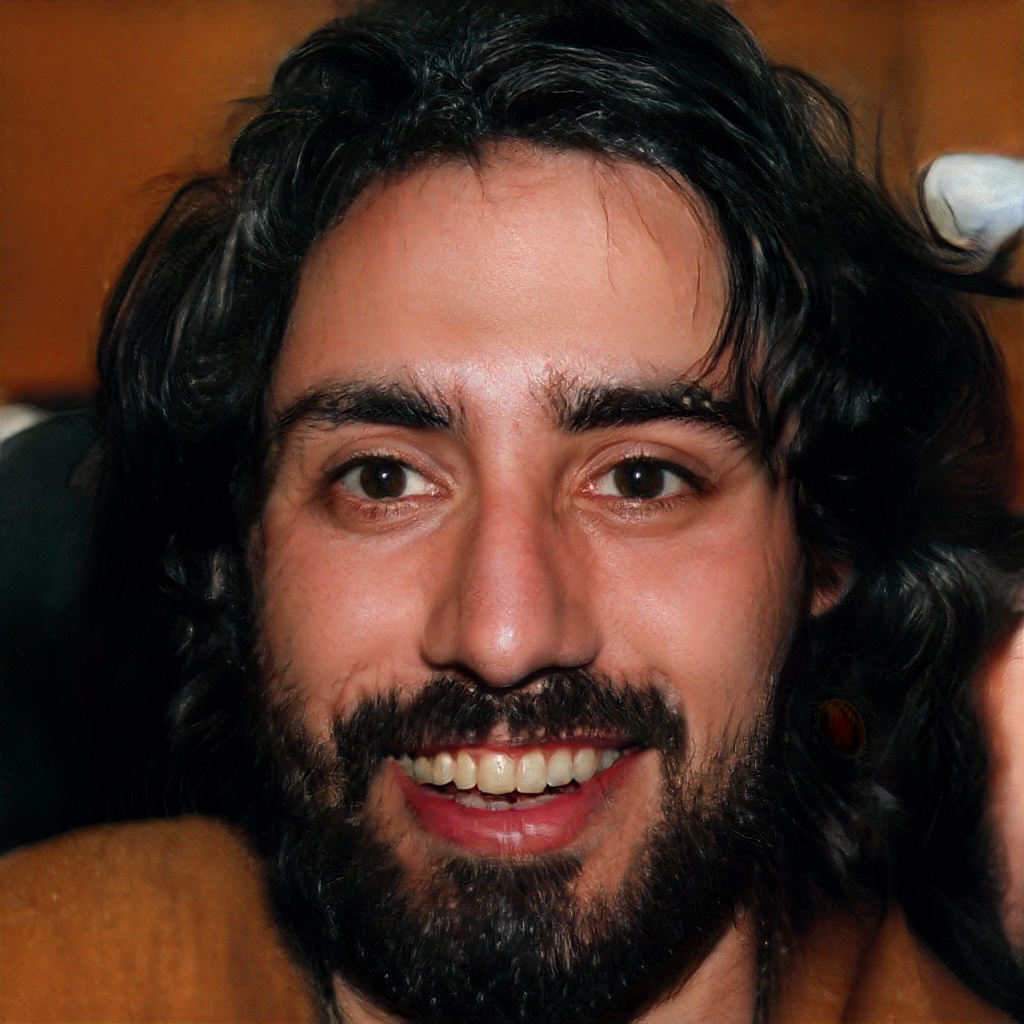}\vv\\
\makebox[\hh][c]{}\hspace{0.5mm}%
\renewcommand{\seed}{seed829}
\includegraphics[height=\h]{figures/Stylemix/\seed.\extt}\hfill%
\includegraphics[height=\h]{figures/Stylemix/\seed-coarse-\varA.\extt}%
\includegraphics[height=\h]{figures/Stylemix/\seed-coarse-\varB.\extt}%
\includegraphics[height=\h]{figures/Stylemix/\seed-coarse-\varC.\extt}%
\includegraphics[height=\h]{figures/Stylemix/\seed-coarse-\varD.\extt}%
\includegraphics[height=\h]{figures/Stylemix/\seed-coarse-\varE.\extt}\vv\\
\makebox[\hh][c]{}\hspace{0.5mm}%
\renewcommand{\seed}{seed1898}
\includegraphics[height=\h]{figures/Stylemix/\seed.\extt}\hfill%
\includegraphics[height=\h]{figures/Stylemix/\seed-coarse-\varA.\extt}%
\includegraphics[height=\h]{figures/Stylemix/\seed-coarse-\varB.\extt}%
\includegraphics[height=\h]{figures/Stylemix/\seed-coarse-\varC.\extt}%
\includegraphics[height=\h]{figures/Stylemix/\seed-coarse-\varD.\extt}%
\includegraphics[height=\h]{figures/Stylemix/\seed-coarse-\varE.\extt}\vspace*{-3.5mm}\\
\begin{tikzpicture}\draw (0,0) -- (\linewidth,0);\end{tikzpicture}\vspace*{-0.5mm}\\%
\makebox[\hh]{\rotatebox[origin=l]{90}{\makebox[\h][c]{\hspace{-\h}\normalsize{Middle styles from source B}}}}\hspace{0.5mm}%
\renewcommand{\seed}{seed1733}
\includegraphics[height=\h]{figures/Stylemix/\seed.\extt}\hfill%
\includegraphics[height=\h]{figures/Stylemix/\seed-middle-\varA.\extt}%
\includegraphics[height=\h]{figures/Stylemix/\seed-middle-\varB.\extt}%
\includegraphics[height=\h]{figures/Stylemix/\seed-middle-\varC.\extt}%
\includegraphics[height=\h]{figures/Stylemix/\seed-middle-\varD.\extt}%
\includegraphics[height=\h]{figures/Stylemix/\seed-middle-\varE.\extt}\vv\\
\makebox[\hh][c]{}\hspace{0.5mm}%
\renewcommand{\seed}{seed1614}
\includegraphics[height=\h]{figures/Stylemix/\seed.\extt}\hfill%
\includegraphics[height=\h]{figures/Stylemix/\seed-middle-\varA.\extt}%
\includegraphics[height=\h]{figures/Stylemix/\seed-middle-\varB.\extt}%
\includegraphics[height=\h]{figures/Stylemix/\seed-middle-\varC.\extt}%
\includegraphics[height=\h]{figures/Stylemix/\seed-middle-\varD.\extt}%
\includegraphics[height=\h]{figures/Stylemix/\seed-middle-\varE.\extt}\vspace*{-3.5mm}\\
\begin{tikzpicture}\draw (0,0) -- (\linewidth,0);\end{tikzpicture}\vspace*{-0.5mm}\\%
\makebox[\hh]{\rotatebox[origin=l]{90}{\makebox[\h][c]{\hspace{-1mm}\normalsize{Fine from B}}}}\hspace{0.5mm}%
\renewcommand{\seed}{seed845}
\includegraphics[height=\h]{figures/Stylemix/\seed.\extt}\hfill%
\raisebox{0mm}[0mm][0mm]{\makebox[0mm]{\hspace*{-1.5mm}\begin{tikzpicture}\draw (0,0mm) -- (0,-196.2mm);\end{tikzpicture}}}%
\includegraphics[height=\h]{figures/Stylemix/\seed-fine-\varA.\extt}%
\includegraphics[height=\h]{figures/Stylemix/\seed-fine-\varB.\extt}%
\includegraphics[height=\h]{figures/Stylemix/\seed-fine-\varC.\extt}%
\includegraphics[height=\h]{figures/Stylemix/\seed-fine-\varD.\extt}%
\includegraphics[height=\h]{figures/Stylemix/\seed-fine-\varE.\extt}\\
\end{minipage}
\caption{\label{fig:stylemix}%
Two sets of images were generated from their respective latent codes (sources A and B); the rest of the images were generated by copying a specified subset of styles from source B and taking the rest from source A. 
Copying the styles corresponding to coarse spatial resolutions ($4^2$ -- $8^2$) brings high-level aspects such as pose, general hair style, face shape, and eyeglasses from source B, while all colors (eyes, hair, lighting) and finer facial features resemble A.
If we instead copy the styles of middle resolutions ($16^2$ -- $32^2$) from B, we inherit smaller scale facial features, hair style, eyes open/closed from B, while the pose, general face shape, and eyeglasses from A are preserved.
Finally, copying the fine styles ($64^2$ -- $1024^2$) from B brings mainly the color scheme and microstructure.
\vspace*{-1mm}
}
\end{figure*}
}

\newcommand{\fignoisedetail}{
\renewcommand{\h}{27.5mm}
\begin{figure}[t]
\includegraphics[width=\h]{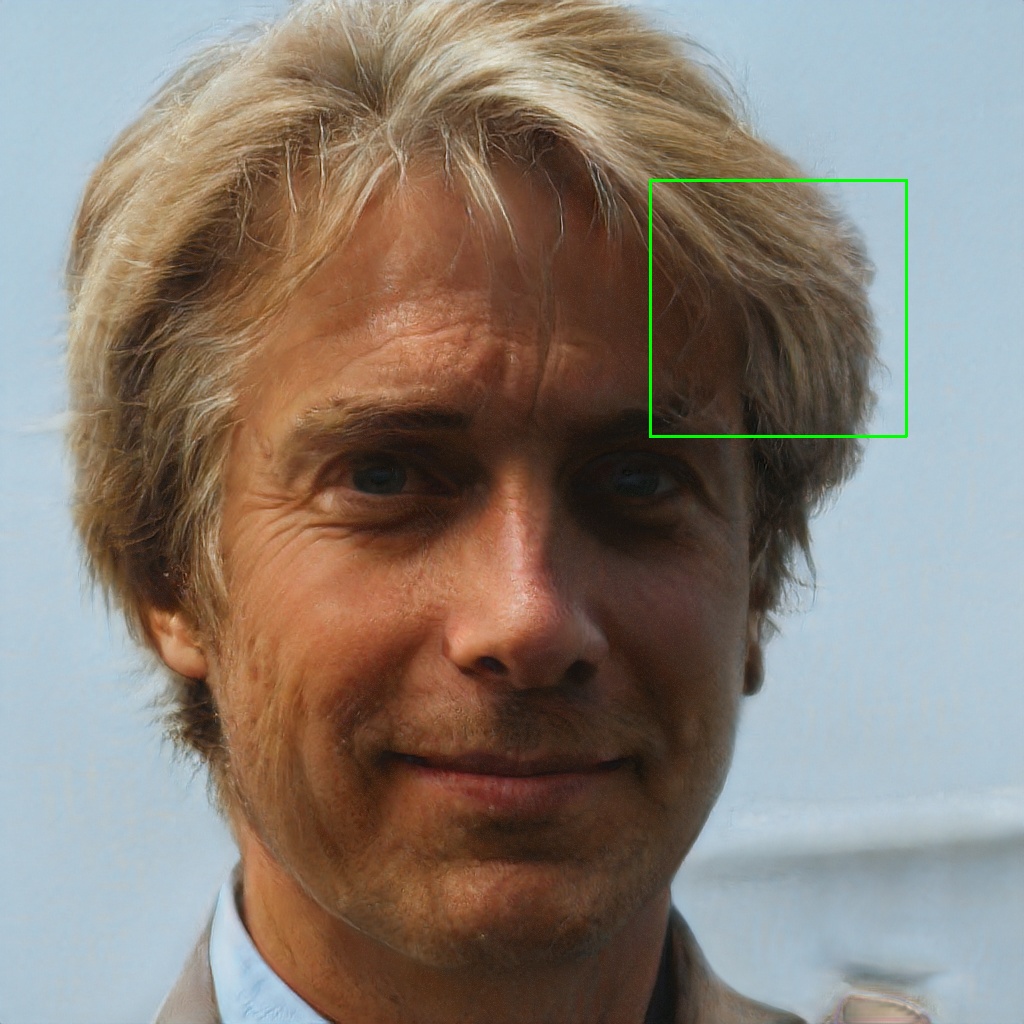}\hfill%
\includegraphics[width=\h]{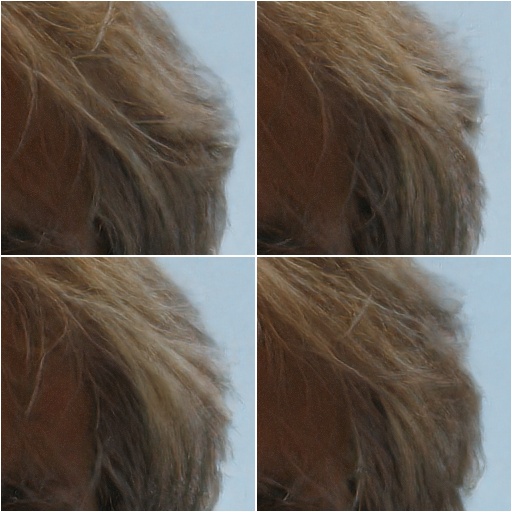}\hfill%
\includegraphics[width=\h]{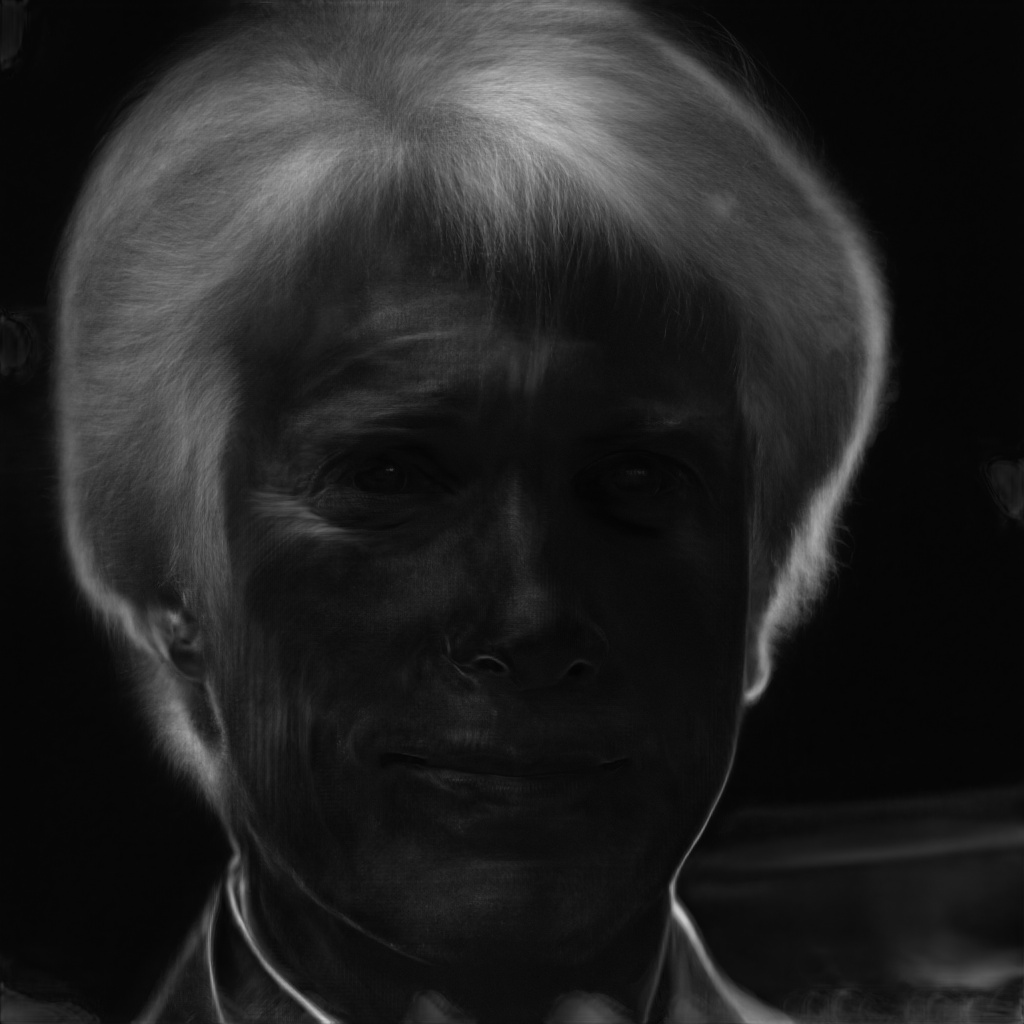}\\
\includegraphics[width=\h]{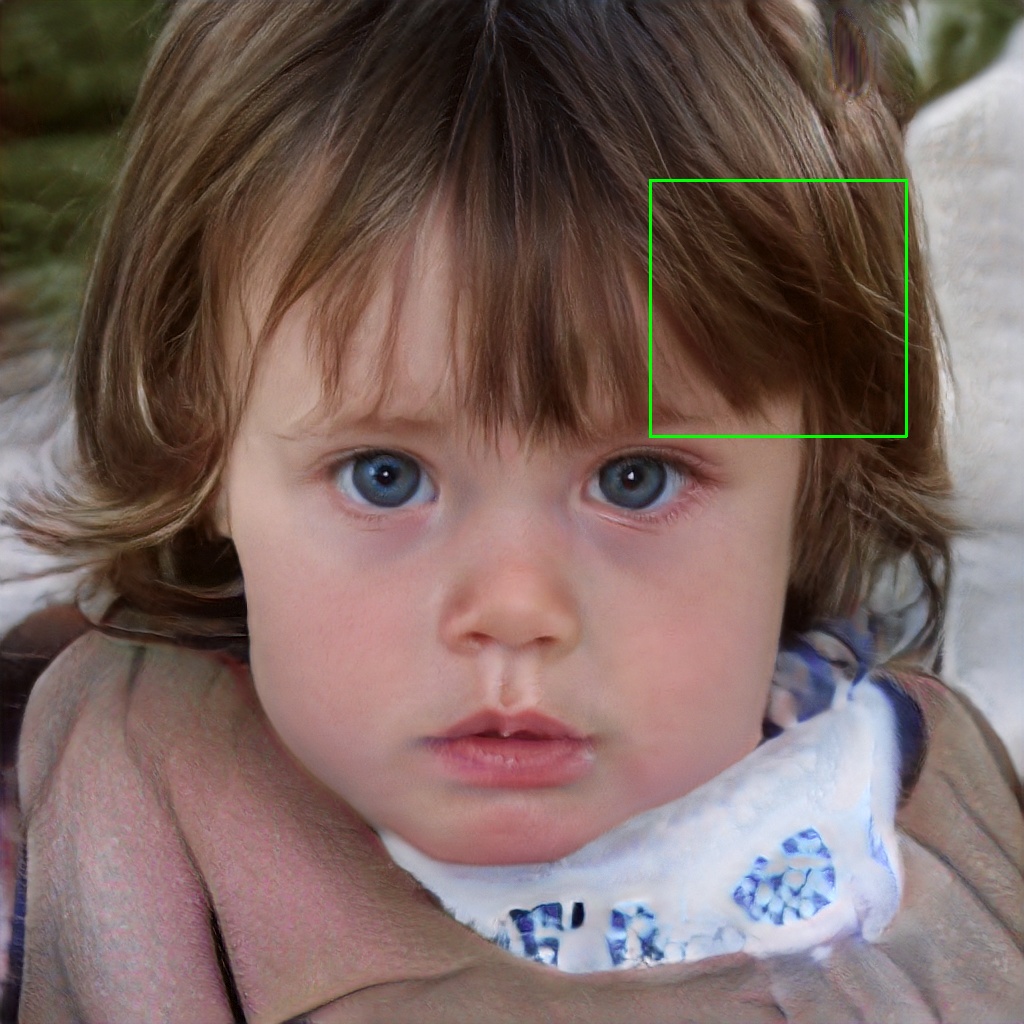}\hfill%
\includegraphics[width=\h]{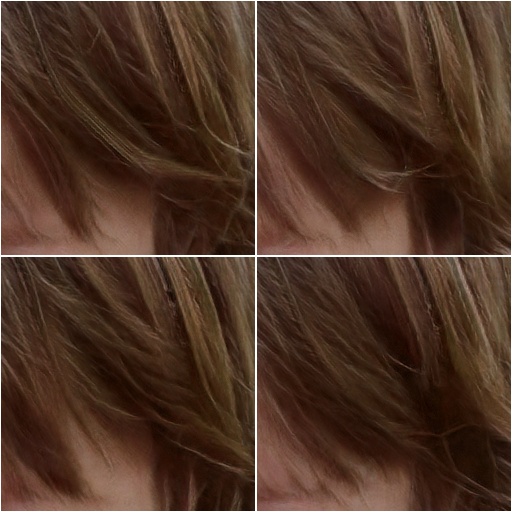}\hfill%
\includegraphics[width=\h]{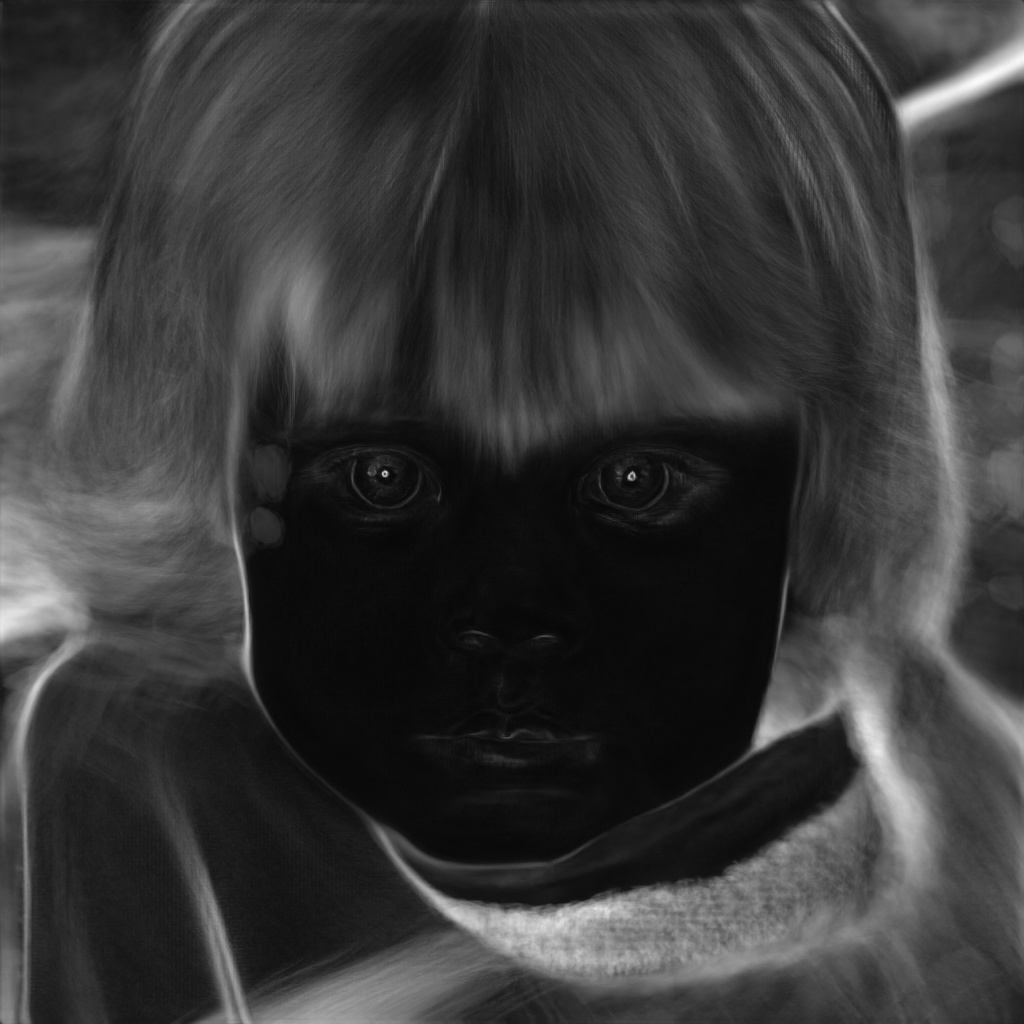}\\
\footnotesize{%
\makebox[\h][c]{(a) Generated image}\hfill%
\makebox[\h][c]{(b) Stochastic variation}\hfill%
\makebox[\h][c]{(c) Standard deviation}}\\\vspace{-1mm}
\caption{Examples of stochastic variation. (a) Two generated images. (b) Zoom-in with different realizations of input noise. While the overall appearance is almost identical, individual hairs are placed very differently. (c) Standard deviation of each pixel over 100 different realizations, highlighting which parts of the images are affected by the noise. The main areas are the hair, silhouettes, and parts of background, but there is also interesting stochastic variation in the eye reflections. Global aspects such as identity and pose are unaffected by stochastic variation.
}\vspace{-1mm}
\label{fig:noisedetail}
\end{figure}
}

\newcommand{\fignoisemix}{
\renewcommand{\h}{0.245\linewidth}
\renewcommand{\hh}{4mm}
\renewcommand{\hhh}{8.83mm}
\begin{figure}[t]
\includegraphics[width=\h]{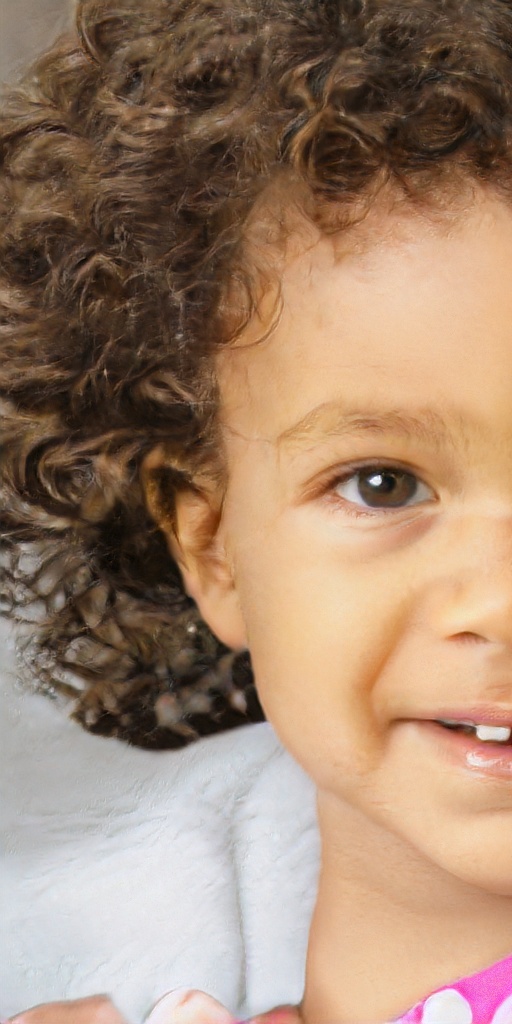}\hspace*{0.35mm}%
\includegraphics[width=\h]{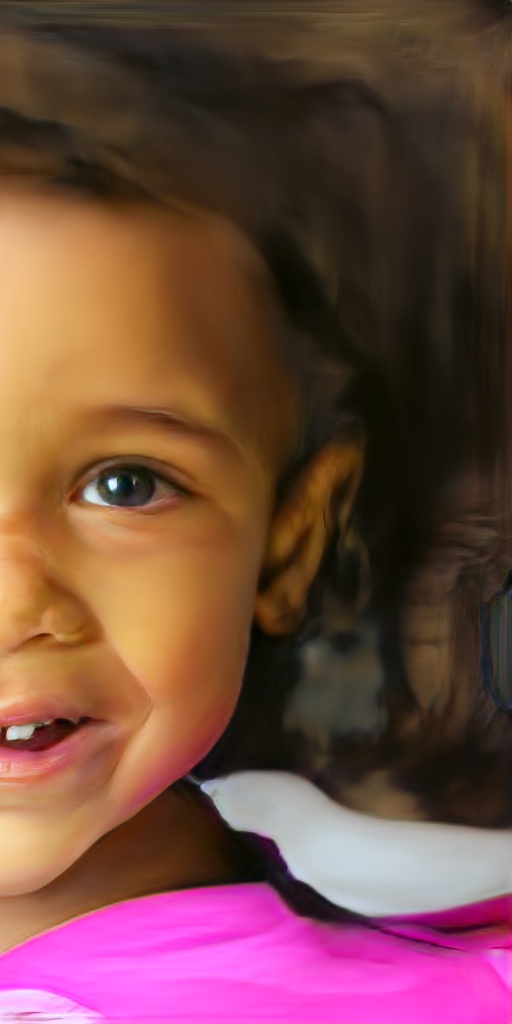}\hfill%
\includegraphics[width=\h]{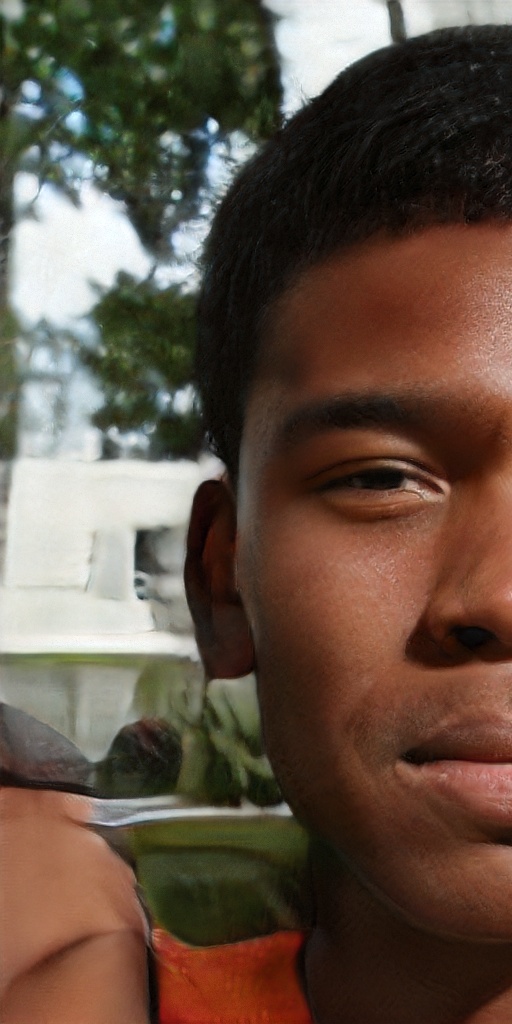}\hspace*{0.35mm}%
\includegraphics[width=\h]{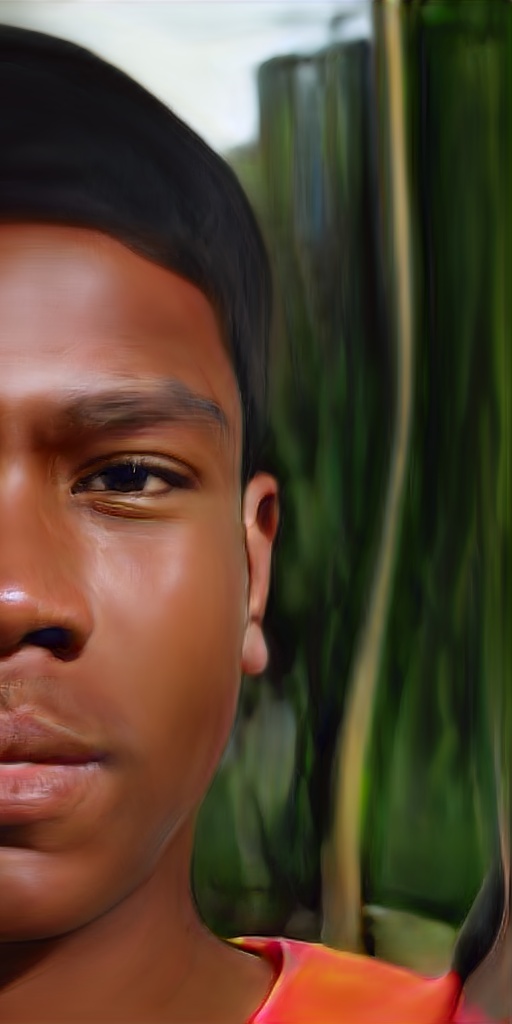}\vspace*{-\baselineskip}\\
\raisebox{1mm}[0mm][0mm]{\makebox[\h][c]{(a)}\hspace*{0.35mm}\makebox[\h][c]{(b)}}\\
\includegraphics[width=\h]{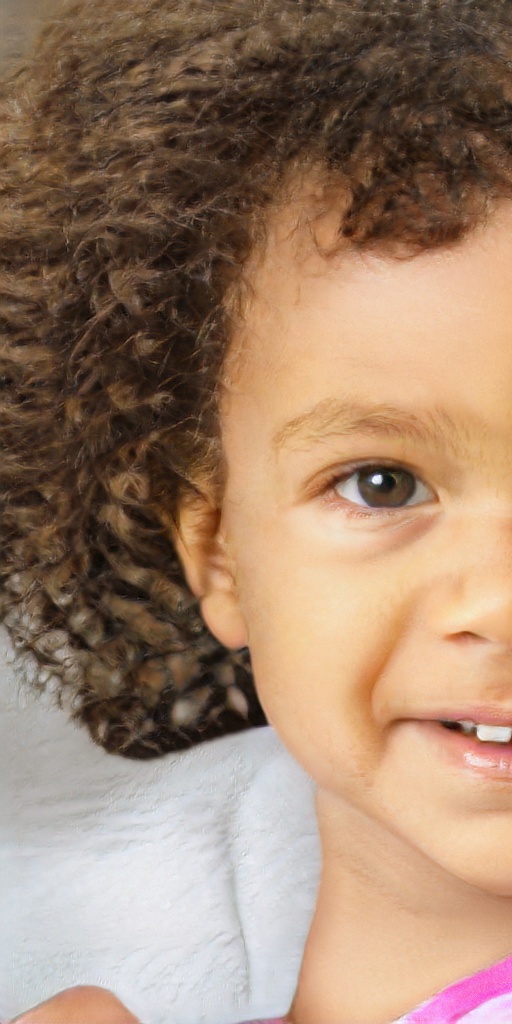}\hspace*{0.35mm}%
\includegraphics[width=\h]{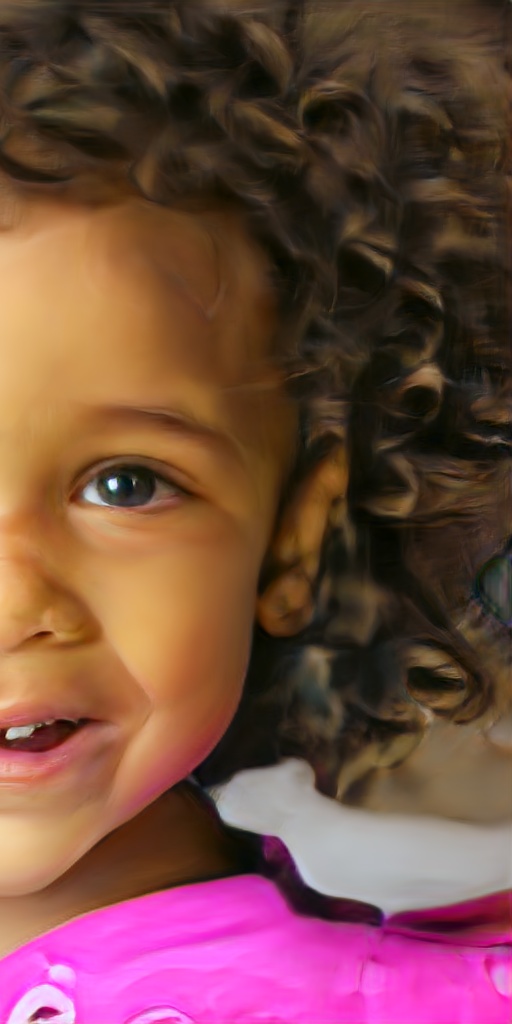}\hfill%
\includegraphics[width=\h]{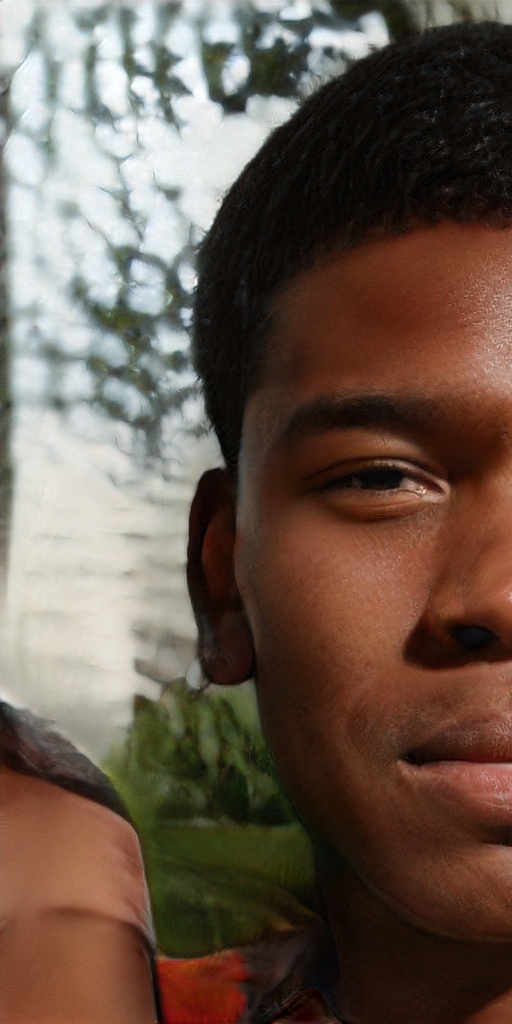}\hspace*{0.35mm}%
\includegraphics[width=\h]{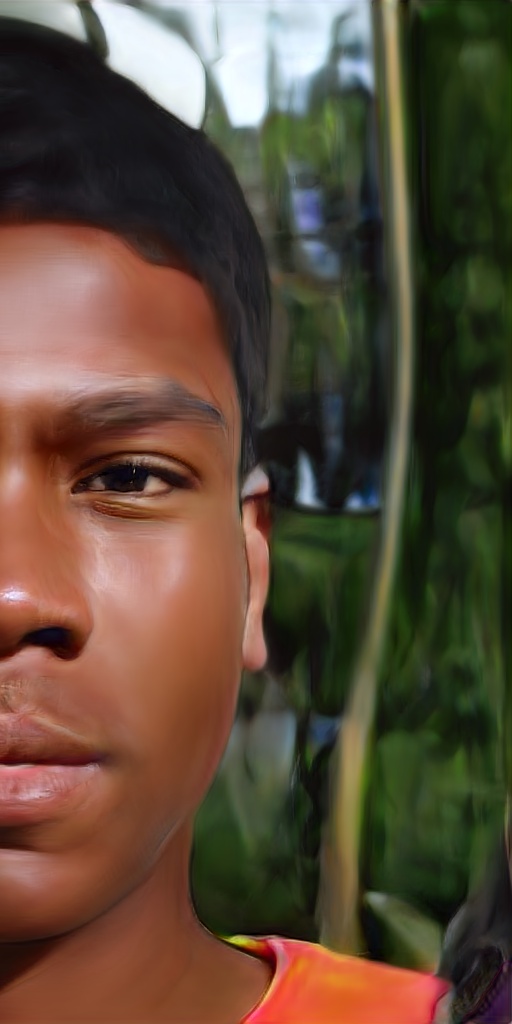}\vspace*{-\baselineskip}\\
\raisebox{1mm}[0mm][0mm]{\makebox[\h][c]{(c)}\hspace*{0.35mm}\makebox[\h][c]{(d)}}\\
\caption{Effect of noise inputs at different layers of our generator. (a) Noise is applied to all layers. (b) No noise. (c) Noise in fine layers only ($64^2$ -- $1024^2$). (d) Noise in coarse layers only \mbox{($4^2$ -- $32^2$)}. We can see that the artificial omission of noise leads to featureless ``painterly'' look. Coarse noise causes large-scale curling of hair and appearance of larger background features, while the fine noise brings out the finer curls of hair, finer background detail, and skin pores.
}
\label{fig:noisemix}
\end{figure}
}

\newcommand{\figquality}{
\begin{figure}[t]
\includegraphics[width=\linewidth]{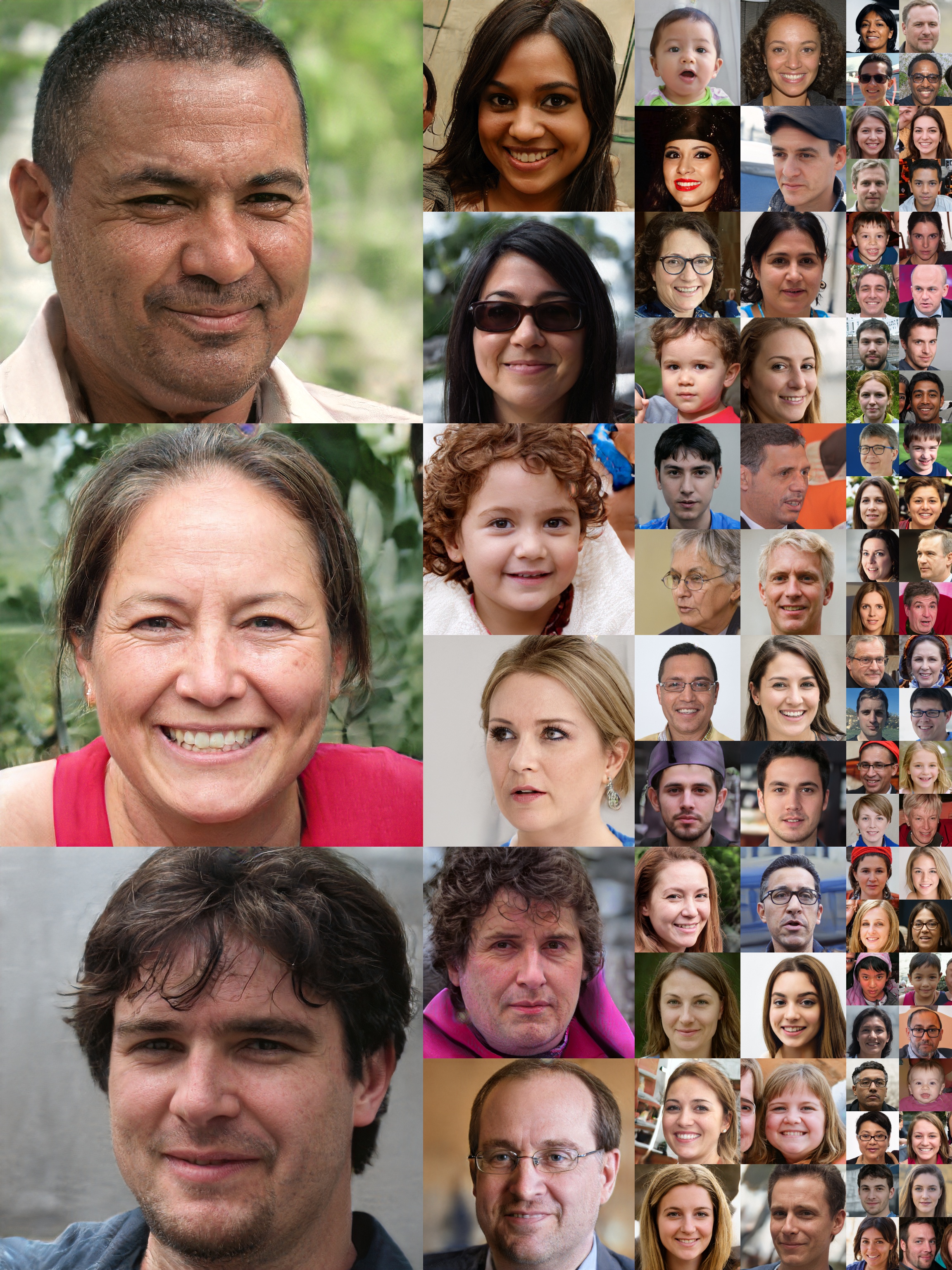}
\caption{Uncurated set of images produced by our style-based generator (config \arch{f}) with the FFHQ dataset. Here we used a variation of the truncation \FINAL{trick \cite{Marchesi2017, Brock2018, Kingma2018}} with $\psi=0.7$ for resolutions $4^2-32^2$. Please see the accompanying video for more results.
}
\label{fig:quality}
\vspace{-2mm}
\end{figure}
}

\newcommand{\figqualitybedroom}{
\begin{figure}[t]
\includegraphics[width=\linewidth]{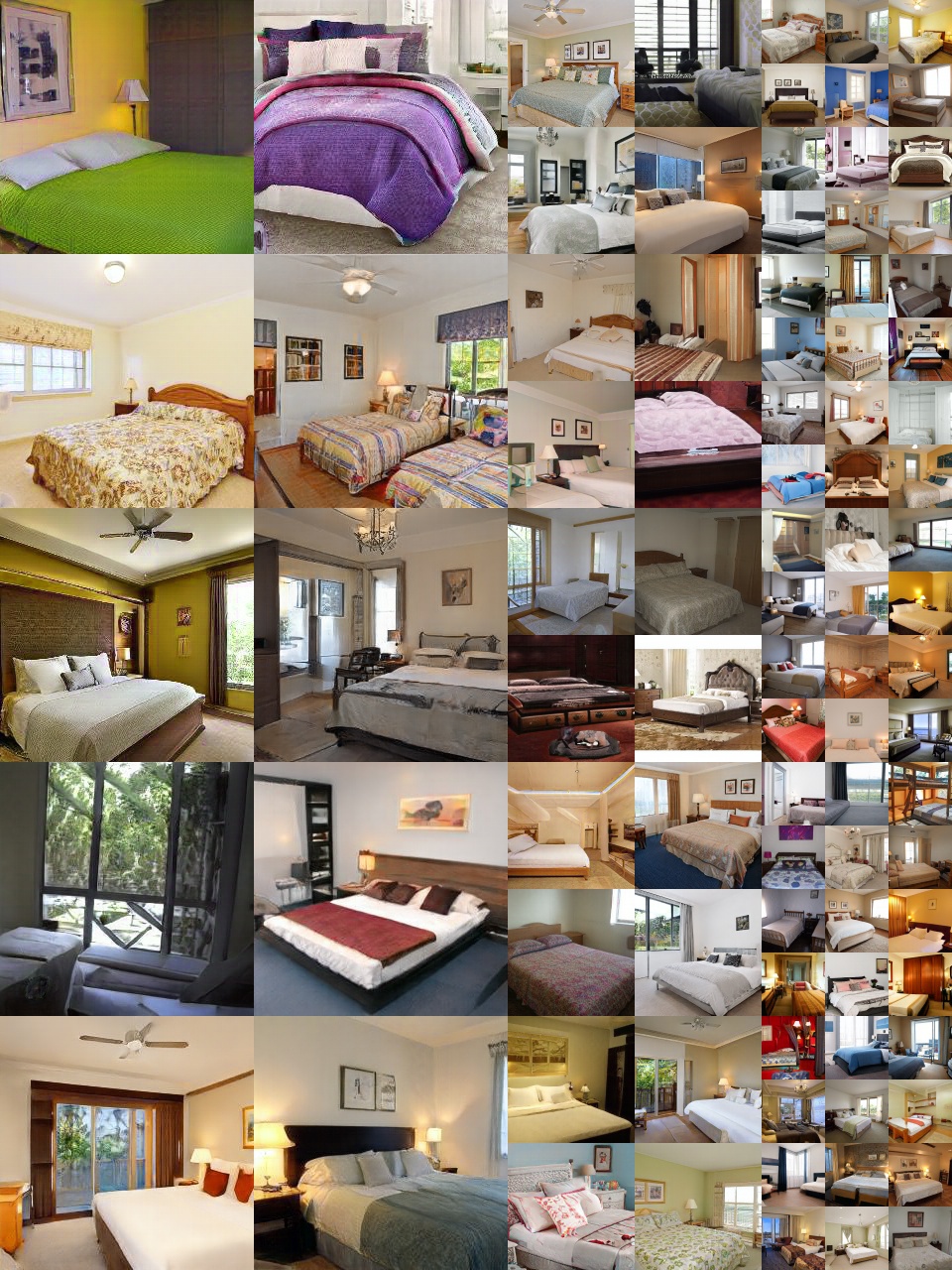}
\caption{Uncurated set of images produced by our style-based generator (config \arch{f}) with the \textsc{LSUN Bedroom} dataset at $256^2$. FID computed for 50K images was 2.65. %
}
\label{fig:qualitybedroom}
\vspace{-2mm}
\end{figure}
}

\newcommand{\figqualitycar}{
\begin{figure}[t]
\includegraphics[width=\linewidth]{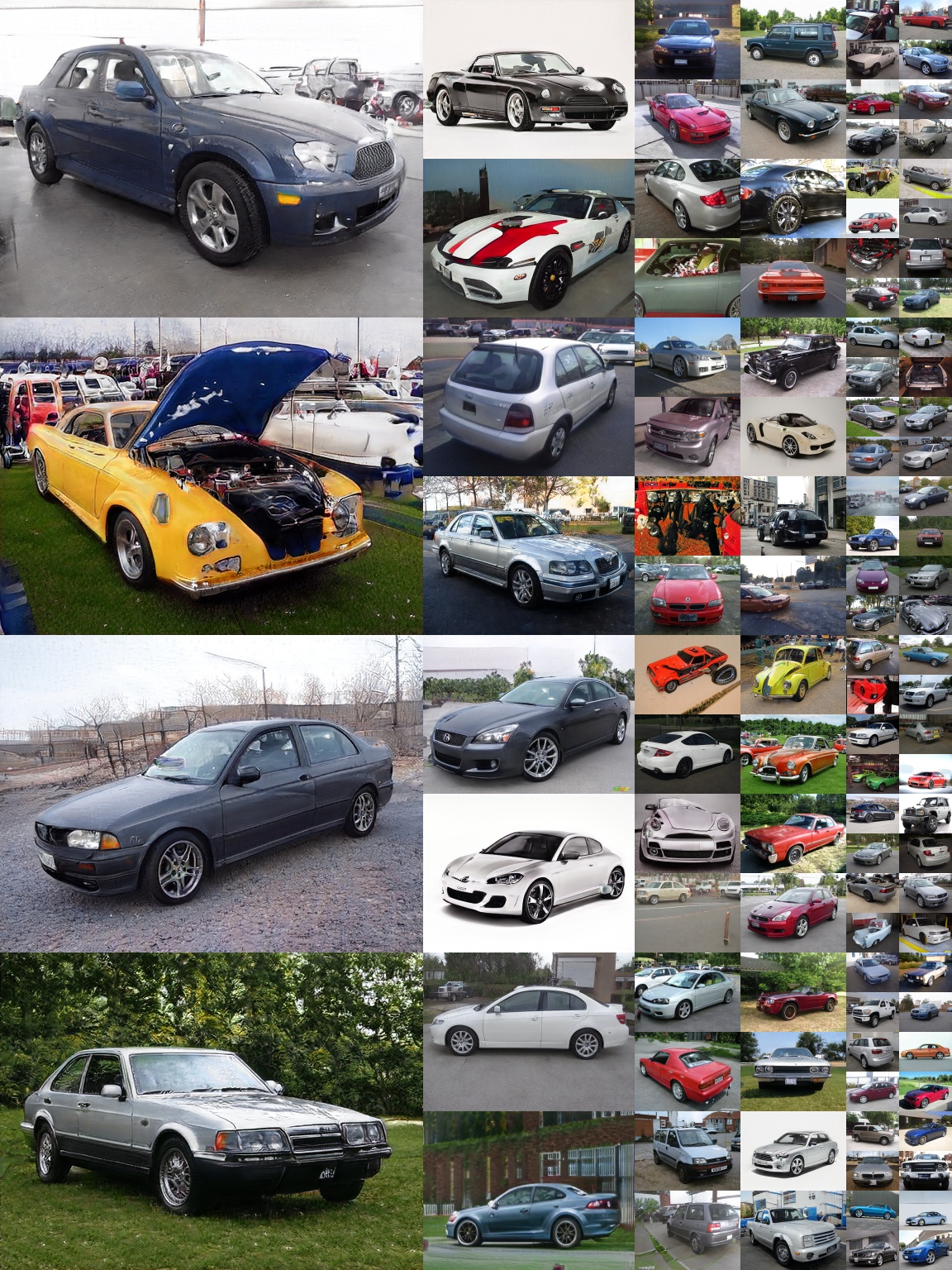}
\caption{Uncurated set of images produced by our style-based generator (config \arch{f}) with the \textsc{LSUN Car} dataset at $512 \times 384$. FID computed for 50K images was 3.27. %
}
\label{fig:qualitycar}
\vspace{0mm}
\end{figure}
}

\newcommand{\figqualitycat}{
\begin{figure}[t]
\includegraphics[width=\linewidth]{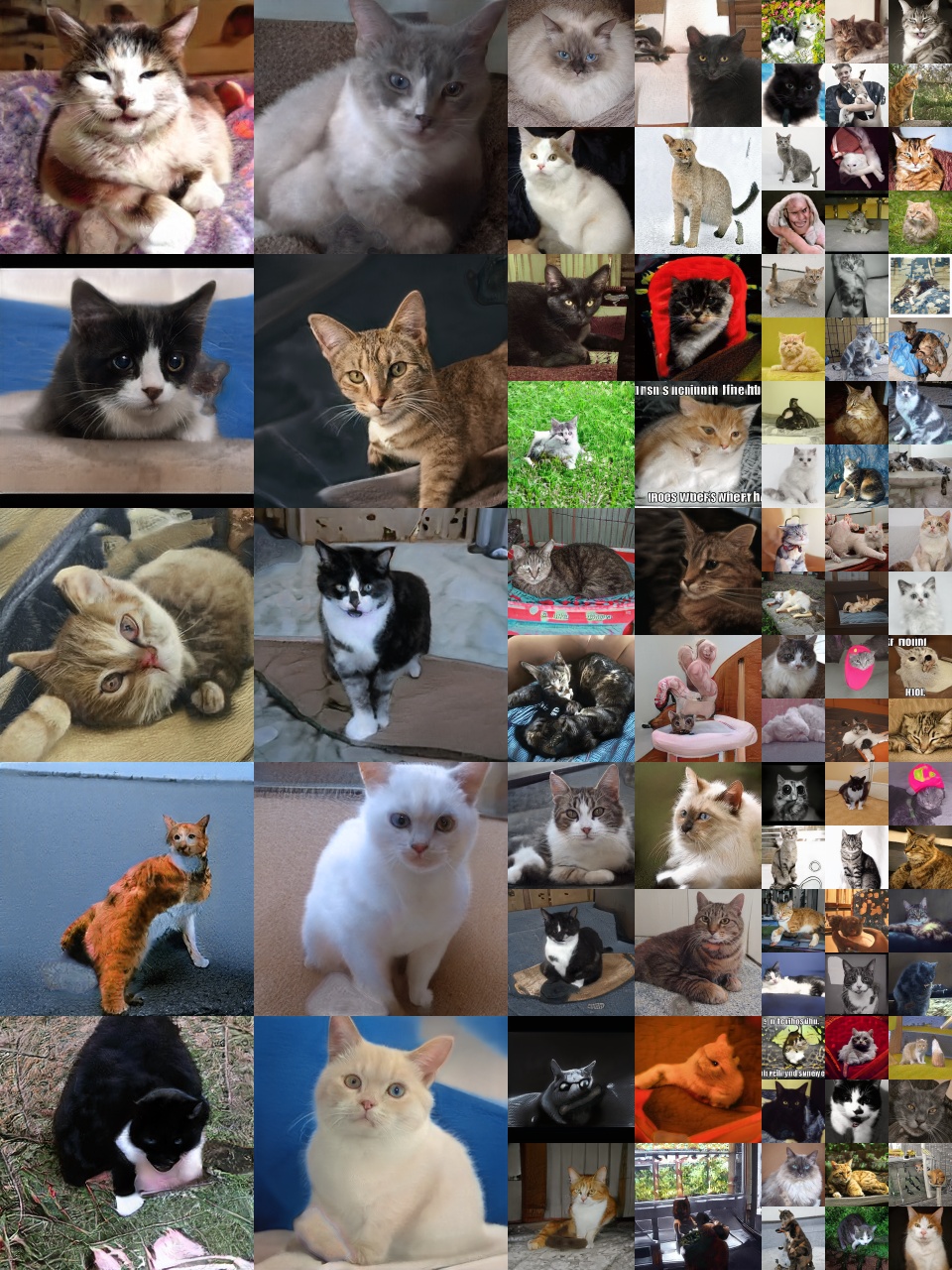}
\caption{Uncurated set of images produced by our style-based generator (config \arch{f}) with the \textsc{LSUN Cat} dataset at $256^2$. FID computed for 50K images was 8.53. %
}
\label{fig:qualitycat}
\vspace{-2mm}
\end{figure}
}

\newcommand{\figillustration}{
\begin{figure}
\centering
\includegraphics[width=0.97\linewidth]{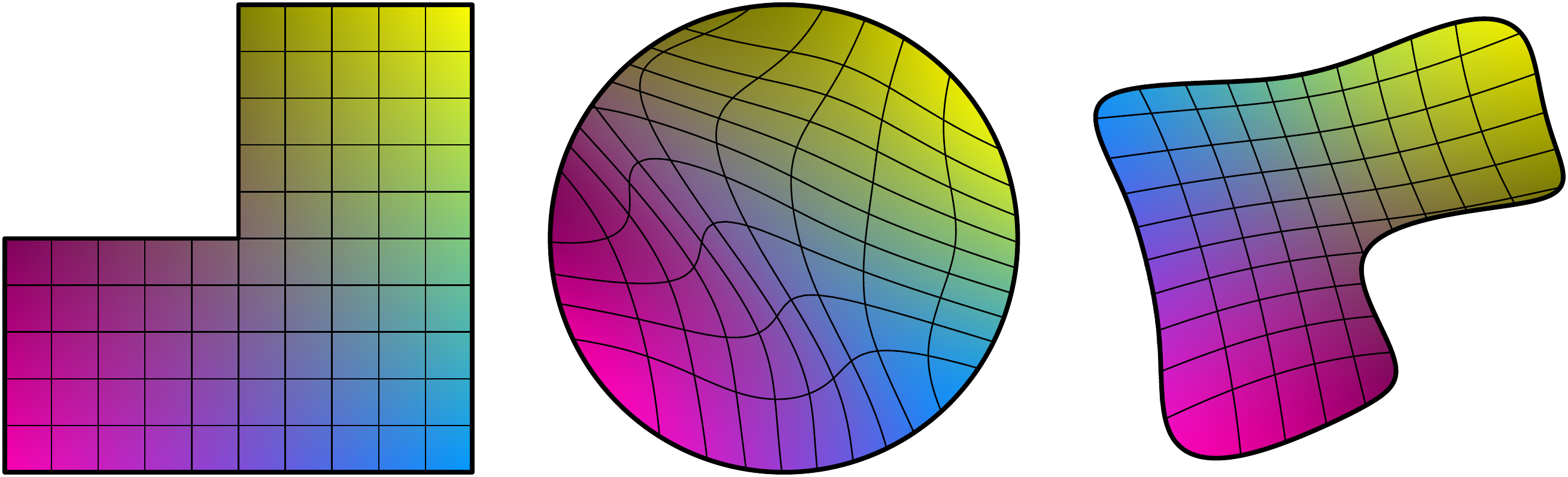}\\
\footnotesize
\makebox[0.33\linewidth][c]{(a) Distribution of}\hfill%
\makebox[0.33\linewidth][c]{(b) Mapping from}\hfill%
\makebox[0.33\linewidth][c]{(c) Mapping from}\\
\makebox[0.33\linewidth][c]{features in training set}\hfill%
\makebox[0.33\linewidth][c]{$\ZZ$ to features}\hfill%
\makebox[0.33\linewidth][c]{$\WW$ to features}\vspace{1.5mm}\\
\caption{%
Illustrative example with two factors of variation (image features, e.g.,~masculinity and hair length).
(a) An example training set where some combination (e.g.,~long haired males) is missing.
(b) This forces the mapping from $\ZZ$ to image features to become curved so that the forbidden combination disappears in $\ZZ$ to prevent the sampling of invalid combinations.
(c) The learned mapping from $\ZZ$ to $\WW$ is able to ``undo'' much of the warping.
}
\label{fig:illustration}
\end{figure}
}

%% file: supplemental-body.tex
\ifarxiv
    \newcommand{\reftabFID}{Table~\ref{tab:FID}}
    \newcommand{\shortreftabFID}{Table~\ref{tab:FID}}
    \newcommand{\refsecseparability}{Section~\ref{sec:separability}}
    \newcommand{\refsectruncation}{Appendix~\ref{sec:truncation}}
\else
    \newcommand{\reftabFID}{Table~1 of the paper}
    \newcommand{\shortreftabFID}{Table~1}
    \newcommand{\refsecseparability}{Section~4.2 of the paper}
    \newcommand{\refsectruncation}{Appendix~B}
\fi

\section{Hyperparameters and training details}
\label{sec:hyperparams}

We build upon the official TensorFlow~\cite{Tensorflow} implementation of Progressive GANs by Karras~et~al.~\cite{Karras2017}, from which we inherit most of the training details.\footnote{\texttt{\scriptsize https://github.com/tkarras/progressive\_growing\_of\_gans}} This original setup corresponds to configuration \arch{a} in \reftabFID.
In particular, we use the same \FINAL{discriminator architecture}, resolution-dependent minibatch sizes, Adam~\cite{Adam} hyperparameters, \FINAL{and exponential moving average of the generator}. We enable mirror augmentation for CelebA-HQ and FFHQ, \FINAL{but disable it for LSUN}.
Our training time is approximately one week on an NVIDIA DGX-1 with 8 Tesla V100 GPUs.

For our improved baseline (\arch{b} in \shortreftabFID), we make several modifications to improve the overall result quality.
We replace the nearest-neighbor up/downsampling in both networks with bilinear sampling, which we implement by low-pass filtering the activations with a separable 2\textsuperscript{nd} order binomial filter after each upsampling layer and before each downsampling \FINAL{layer \cite{zhang2019}}.
We implement progressive growing the same way as Karras~et~al.~\cite{Karras2017}, but we start from $8^2$ images instead of $4^2$.
For the FFHQ dataset, we switch from WGAN-GP to the non-saturating loss \cite{Goodfellow2014} with $R_1$ regularization \cite{Mescheder2018} using $\gamma=10$.
With $R_1$ we found that the FID scores keep decreasing for considerably longer than with WGAN-GP, and we thus increase the training time from 12M to 25M images.
We use the same learning rates as Karras~et~al.~\cite{Karras2017} for FFHQ, but we found that setting the learning rate to 0.002 instead of 0.003 for $512^2$ and $1024^2$ leads to better stability with CelebA-HQ.

For our style-based generator (\arch{f} in \shortreftabFID), we use leaky ReLU \cite{Maas2013} with $\alpha=0.2$ and equalized learning rate~\cite{Karras2017} for all layers.
We use the same feature map counts in our convolution layers as Karras~et~al.~\cite{Karras2017}.
Our mapping network consists of 8 fully-connected layers, and the dimensionality of all input and output activations\,---\,including $\zz$ and $\ww$\,---\,is 512.
We found that increasing the depth of the mapping network tends to make the training unstable with high learning rates.
We thus reduce the learning rate by two orders of magnitude for the mapping network, i.e., $\lambda' = 0.01\cdot\lambda$.
We initialize all weights of the convolutional, fully-connected, and affine transform layers using $\mathcal{N}(0,1)$.
The constant input in synthesis network is initialized to one.
The biases and noise scaling factors are initialized to zero, except for the biases associated with $\yy_s$ that we initialize to one.

The classifiers used by our separability metric (\refsecseparability) have the same architecture as our discriminator except that minibatch standard deviation~\cite{Karras2017} is disabled.
We use the learning rate of $10^{-3}$, minibatch size of 8, Adam optimizer, and training length of 150,000 images.
The classifiers are trained independently of generators, and the same~40 classifiers, one for each CelebA attribute, are used for measuring the separability metric for all generators.
We will release the pre-trained classifier networks so that our measurements can be reproduced.

\figpathlenbehavior

We do not use batch normalization \cite{Ioffe2015}, spectral normalization \cite{Miyato2018B}, attention mechanisms \cite{Zhang2018sagan}, dropout \cite{srivastava2014}, or pixelwise feature vector normalization \cite{Karras2017} in our networks.

\section{Training convergence}
\figqualitybedroom

Figure~\ref{fig:pathlenbehavior} shows how the FID and perceptual path length metrics evolve during the training of our configurations \arch{b} and \arch{f} with the FFHQ dataset.
With $R_1$ regularization active in both configurations, FID continues to slowly decrease as the training progresses, 
motivating our choice to increase the training time from 12M images to 25M images.
Even when the training has reached the full $1024^2$ resolution, the slowly rising path lengths indicate that the improvements in FID come at the cost of a more entangled representation.
Considering future work, it is an interesting question whether this is unavoidable, or if it were possible to encourage shorter path lengths without compromising the convergence of FID.

\section{Other datasets}
\label{sec:otherdatasets}

Figures~\ref{fig:qualitybedroom},~\ref{fig:qualitycar}, and~\ref{fig:qualitycat} show an uncurated set of results for \textsc{LSUN} \cite{LSUN} \textsc{Bedroom}, \textsc{Cars}, and \textsc{Cats}, respectively. In these images we used the truncation trick from \refsectruncation with $\psi=0.7$ for resolutions $4^2-32^2$. The accompanying video provides results for style mixing and stochastic variation tests. As can be seen therein, in case of \textsc{Bedroom} the coarse styles basically control the viewpoint of the camera, middle styles select the particular furniture, and fine styles deal with colors and smaller details of materials. In \textsc{Cars} the effects are roughly similar. Stochastic variation affects primarily the fabrics in \textsc{Bedroom}, backgrounds and headlamps in \textsc{Cars}, and fur, background, and interestingly, the positioning of paws in \textsc{Cats}. Somewhat surprisingly the wheels of a car never seem to rotate based on stochastic inputs.

These datasets were trained using the same setup as FFHQ for the duration of 70M images for \textsc{Bedroom} and \textsc{Cats}, and 46M for \textsc{Cars}. We suspect that the results for \textsc{Bedroom} are starting to  approach the limits of the training data, as in many images the most objectionable issues are the severe compression artifacts that have been inherited from the low-quality training data. \textsc{Cars} has much higher quality training data that also allows higher spatial resolution ($512\times384$ instead of $256^2$), and \textsc{Cats} continues to be a difficult dataset due to the high intrinsic variation in poses, zoom levels, and backgrounds.

\figqualitycar
\figqualitycat